\documentclass[conference]{IEEEtran}
\IEEEoverridecommandlockouts
    \usepackage{cite}
    \usepackage[bookmarks=false]{hyperref}
    \usepackage[pdftex]{graphicx}
    \usepackage[caption=false,font=footnotesize]{subfig}
    \usepackage{amsmath}
    \usepackage{dblfloatfix}

\begin{document}

\title{Joint Enhancement and Denoising Method \\
via Sequential Decomposition}

\author{\IEEEauthorblockN{Xutong Ren$^1$, Mading Li$^1$, Wen-Huang Cheng$^2$ and Jiaying Liu$^{1,\ast}$
\thanks{\footnotesize{$^{\ast}$Corresponding author \newline
This work was supported by the National Natural Science Foundation of China under Contract 61772043 and the Ministry of Science and Technology of Taiwan under Grants MOST-105-2628-E-001-003-MY3 and MOST-106-3114-E-002-009.}}
}
\IEEEauthorblockA{$^1$Institute of Computer Science and Technology, Peking University, Beijing, China\\
$^2$Research Center for Information Technology Innovation (CITI), Academia Sinica, Taiwan}}

\maketitle

\begin{abstract}

     Many low-light enhancement methods ignore intensive noise in original images. As a result, they often simultaneously enhance the noise as well. Furthermore, extra denoising procedures adopted by most methods may ruin the details. In this paper, we introduce a joint low-light enhancement and denoising strategy, aimed at obtaining great enhanced low-light images while getting rid of the inherent noise issue simultaneously. The proposed method performs Retinex model based decomposition in a successive sequence, which sequentially estimates a piece-wise smoothed illumination and an noise-suppressed reflectance. After getting the illumination and reflectance map, we adjust the illumination layer and generate our enhancement result. In this noise-suppressed sequential decomposition process we enforce the spatial smoothness on each component and skillfully make use of weight matrices to suppress the noise and improve the contrast. Results of extensive experiments demonstrate the effectiveness and practicability of our method. It performs well for a wide variety of images, and achieves better or comparable quality compared with the state-of-the-art methods.

\end{abstract}

\IEEEpeerreviewmaketitle

\section{Introduction}

    \begin{figure*} [ht]
        \centering
        \includegraphics[width=0.9\textwidth,height=0.3\textwidth]{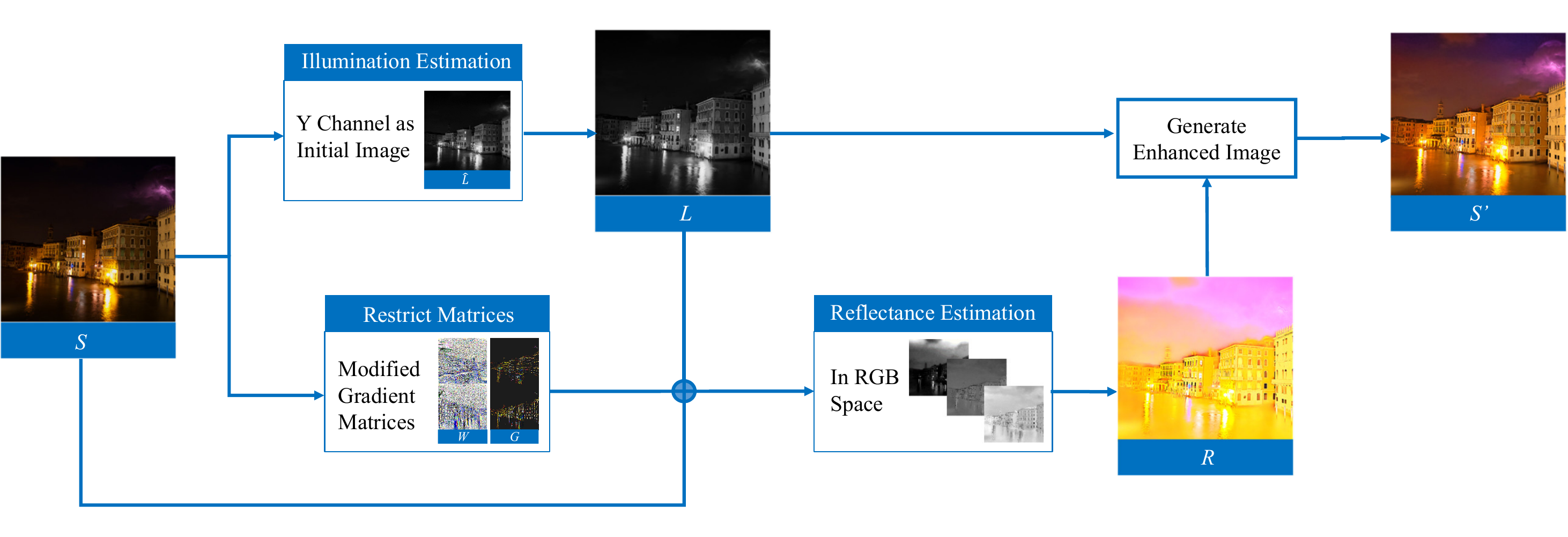}
        \caption{Framework of the proposed method. We first use an initial illumination to refine the final illumination. Following that we use the obtained illumination map and gradient matrices $W$ and $G$ to amplify the reflectance in RGB space. Then the enhanced image is generated from the illumination and reflectance. }
        \label{F_frame}
    \end{figure*}

    As living standard improves rapidly, more and more people nowadays like to go travelling and they often take photos of themselves or the landscape when they are out. And with the booming of social media such as Facebook and YouTube, it nearly becomes a routine for some people to photograph or video their everyday life and share them with others through Internet. However, many photos are captured under low-light circumstance due to backlight, under exposure or dark environment. Although the improvement of skills, equipments, and conditions could to some extent avoid that, it is still inevitable to take photos with undesirable quality, let alone the massive images with the problem acquired in the past. These photos so called low-light images suffer from low definition, low contrast and much noise. Post-processing techniques are required to enhance the visual quality of these images.

    It is the most intuitive and simplest way to directly amplifying the illumination of a low-light image. But this operation results in some other problems like saturating bright areas and losing details. Histogram equalization (HE) based methods \cite{109340} flatten the histogram and stretch the dynamic range of the intensity, alleviating the above problems. But the results of these methods may be under- or over-enhanced with much noise. Some researchers \cite{7351501,7747352} noticed the similarity between haze images and the inverted low-light images. Thus they applied dehazing methods to deal with low-light image. With respect to the intensive noise, Li~\emph{et. al.}~\cite{7351501} tried to eliminate the influence of noise via BM3D \cite{4271520} after enhancement. A joint-bilateral filter is applied in \cite{7747352} to suppress the noise after the enhancement.

    Retinex-based methods consider the scene in human¡¯s eyes as the product of reflectance and illumination layers. Logarithmic transformation can simply the multiplication. However, A recent work \cite{7780673} comes up with the opinion that logarithmic transformation is not appropriate despite being widely adopted. To cover the shortage of logarithmic form they give a weighted variational model which estimate both the reflectance and the illumination. The model shows surprising results but the noise is quite observable in the results, especially when there is much noise in original images. Another work \cite{7782813} pays attention to estimate the strengthened illumination map by attaching a coefficient matrix. Although this method obtains impressive results, it also generates the problem of over-enhancement and losing details in bright areas. Besides, since the unprocessed reflectance contains much noise, the enhanced image often has noise and an extra denoising procedure via BM3D \cite{4271520} is also needed. Yue \emph{et. al.} \cite{7924358} concentrated on intrinsic image decomposition and introduces constraints on both reflectance and illumination layers. But similarly, they do not take noise as a component or an influence factor in the decomposition procedure.

    In our work, we consider noise as a non-negligible factor in Retinex based decomposition. Thus low-light enhancement should be aware that eliminating noise must be simultaneously proceeded while enhancing the illumination, not in a separate way. Based on that, we propose an integrative method to simultaneously enhance the images and suppress the noise. Our method estimates both the illumination and the reflectance but in a successive sequence. That is, we first estimate the illumination map, independent from the reflectance map. Then we refine the reflectance on the basis of both the refined illumination and the original image. Given that noise exists in the source image, after we extract the smooth illumination map, most noise is left in the reflectance. Therefore, we use weighted matrices to restrict noise when embellishing the reflectance. We argue that estimating the illumination and reflectance simultaneously using iterative method may introduce more noise to the expected illumination map because noise is often observed in the reflectance image in the classic Retinex decomposition. So a sequential estimation method can obtained a more purified illumination and as a consequence, a better reflectance with noise limited to the minimum. After obtaining the preferred illumination and reflectance, the final enhancement result is generated by combining the reflectance and the Gamma corrected illumination.

    The rest of this paper is organized as follows. The proposed approach is elaborated in Section II. Experimental results are presented in Section III. Finally we draw a conclusion in Section IV.

\section{The Joint Low-Light Enhancement and Denoising Method}

\subsection{Overview}

    The classic Retinex model decomposes images into reflectance and illumination as $S = R \circ L$, where $S$ is the observed image, $R$ and $L$ represent the reflectance and the illumination of the image. The operator $\circ$ denotes the element-wise multiplication.

    Knowing that low light may introduce much noise to the image and enhancing the picture inevitably intensifies the noise at the same time, we hold the view that the classic Retinex model should be modified with a noise term $N$ as follows:
    \begin{equation}
        S = R \circ L + N.
    \end{equation}

    Many methods focus on the illumination component $L$ and simply take $R' = S/L$ as the obtained reflectance, which actually keeps most unpleasant noise intact in the reflectance image for $R' = R + N/L$.

    Thus, those methods always lead to noisy results and often require an extra denoising procedure. However, this may cause the missing of some critical details in the image. Besides, some methods use term $\|R \circ L - S\|^2_F$ in their equations, and in order to calculate both $R$ and $L$ simultaneously, they iteratively update each variable while regarding the other variables as constants. In other words, $L$ is calculated on the basis of the previous result of $R$ in every iteration. But we argue that, during these procedures the noise, which is often observed in the reflectance image, continuously impairs the expected illumination map $L$.

    For those reasons, we propose a new optimization method here that consider both reflectance $R$ and illumination $L$ but calculate them in a way more respective, as well as seeing noise as one of the affecting factors. Fig. \ref{F_frame} shows the framework of our method.

\subsection{The Sequential Estimation}

    As discussed above, we choose to propose sequential equations to acquire the most undisturbed illumination $L$ and the most preferred reflectance $R$:

    \begin{equation}
        \label{E_L}
        \operatorname*{argmin}\limits_{L} \| L - \hat{L} \|^{2}_{F} + \alpha \| \nabla L \|_{1},
    \end{equation} \vspace{-5mm}
    \begin{equation}
        \label{E_R}
        \operatorname*{argmin}\limits_{R} \| R - S/L \|^{2}_{F} + \beta \|W \circ \nabla R \|^2_F + \omega \| \nabla R - G \|^{2}_{F},
    \end{equation}
    where $\alpha$, $\beta$, and $\omega$ are the coefficients that control the importance of different terms. $\|\cdot\|_F$ and $\|\cdot\|_1$ represent the Frobenius norm and $\ell_1$ norm, respectively. In addition, $\nabla$ is the first order differential operator, $\circ$ denotes the element-wise multiplication, $W$ is a weight matrix related to the observed image $S$, and $G$ is the adjusted gradient of $S$. The role of each term in (\ref{E_L}) and (\ref{E_R}) are interpreted below:

    \begin{itemize}
    \item $\| L - \hat{L} \|^{2}_{F}$ takes care of the fidelity between the initial illumination map $\hat{L}$ and the refined one $L$;
    \item $\| \nabla L \|_1$ corresponds to the total variation sparsity and considers the piece-wise smoothness of the illumination map $L$;
    \item $\| R - S/L \|^{2}_{F}$ constrains the fidelity between the observed image $S$ and the recomposed one $R \circ L$. In other words, $R$ and $R'$;
    \item $\|W \circ \nabla R \|^2_F$ enforces the spatial smoothness on the reflectance $R$ accommodatively;
    \item $\| \nabla R - G \|^{2}_{F}$ minimizes the distance between the gradient of the reflectance $R$ and that of the observed image $S$, so that the contrast of the final result can be strengthened.
    \end{itemize}

    In this paper, we assume that, for color images, three channels share the same illumination map. Therefore, we generally set the initial illumination map $\hat{L}$ as $Y$ channel of the input image.

    With the hope of getting a preferable reflectance $R$ whose gradients are smooth in homogeneous areas while undamaged at edges, we reasonably set $W$ as follows:
    \begin{equation}
        W = \frac{1}{|\nabla S|+\epsilon}.
    \end{equation}

    The modified term $\|W \circ \nabla R \|^2_F$ enforces the spatial smoothness on the reflectance $R$ as well, but the extent at different position of the image is under the control of $W$. This is effective especially when there is intensive, large-scale noise in the original image.

    As for the matrix $G$, it is an adjusted version of $\nabla S$, designed to amplify the reflectance $R$ while restrain noise. The formulation of $G$ is given as follows:

    \begin{equation}
    \begin{split}
        &G = (1+\lambda e^{-|\nabla \hat{S}|/\sigma}) \circ \nabla \hat{S}, \\
        \\
        &\nabla \hat{S} =
        \begin{cases}
            0, &\textrm{if} \  |\nabla S| < \varepsilon, \\
            \nabla S, &\textrm{otherwise},
        \end{cases}
    \end{split}
    \end{equation}
    where $\lambda$ controls the degree of the amplification; $\sigma$ controls the amplification rate of different gradients; $\varepsilon$ is the threshold that filters small gradients.

    By suppressing small gradients first, this equation minimizes the possible noise. And then strengthen the overall gradients with alterable proportions.

    For each observed image, matrix $\hat{L}$, $W$ and $G$ only need to be calculated once.

\subsection{The Solution}

    Inspired by low-light image enhancement via illumination map estimation (LIME) \cite{7782813}, we use the alternative $\sum_x \sum_{d\in \{h,v\}} \frac{( \nabla_d L(x) )^2}{|\nabla_d \hat{L}(x)|+\epsilon }$ to approximate $\| \nabla L \|_1$. As a result, the approximate problem to (\ref{E_L}) can be written as follows:

    \begin{equation}
        \label{E_sqr_L}
        \operatorname*{argmin}\limits_{L} \| L - \hat{L} \|^{2}_{F} + \alpha \sum_x \sum_{d\in \{h,v\}} \frac{( \nabla_d L(x) )^2}{|\nabla_d \hat{L}(x)|+\epsilon }.
    \end{equation}

    This change does not influence the result much because according to the first term $\| L - \hat{L} \|^{2}_{F}$, the gradients of $L$ should also be similar to those of $\hat{L}$. For convenience, we put (\ref{E_sqr_L}) in a simpler from, where $A_d (x)$ represents $\frac{\alpha }{|\nabla_d \hat{L}(x)|+\epsilon }$:

    \begin{equation}
        \label{E_sqr_sim_L}
        \operatorname*{argmin}\limits_{L} \| L - \hat{L} \|^{2}_{F} + \sum_x \sum_{d\in \{h,v\}} A_d(x) \cdot ( \nabla_d L(x) )^2.
    \end{equation}

    As can be observed, (\ref{E_sqr_sim_L}) only involves quadratic terms. Thus, by differentiating (\ref{E_sqr_sim_L}) with respect to $L$ and setting the derivative to 0, the problem can be directly figured out by solving the following:
    \begin{equation}
        \left( I+\sum_{d\in \{h,v\}} D^T_d \textrm{Diag} (a_d) D_d \right) l = \hat{l},
    \end{equation}
    where $I$ is the identity matrix with proper size. And $D$ contains $D_h$ and $D_v$, which are the Toeplitz matrices from the discrete gradient operators with forward difference. Further, $x$ is the vectorized version of $X$ and the operator Diag($x$) is to construct a diagonal matrix using vector $x$. Then we can easily solve it to obtain the evaluated $L$.

    Similarly, by differentiating (\ref{E_R}) with respect to $R$ and setting the derivative to 0, we have the following equation:

    \begin{eqnarray}
    \begin{split}
       &\left( I+ \sum\limits_{d\in\{h,v\}} \beta D^T_d \, \textrm{Diag} \, (w_d) D_d + \sum\limits_{d\in\{h,v\}} \omega D^T_d D_d \right ) r \\
       \\
       &= s/l+\sum\limits_{d\in \{h,v\}} \omega D^T_d g_d,
    \end{split}
    \end{eqnarray}

    \begin{figure} [ht]
        \centering
        \subfloat[] {
            \begin{minipage}[b]{0.1\textwidth}
            \includegraphics[width=1\textwidth]{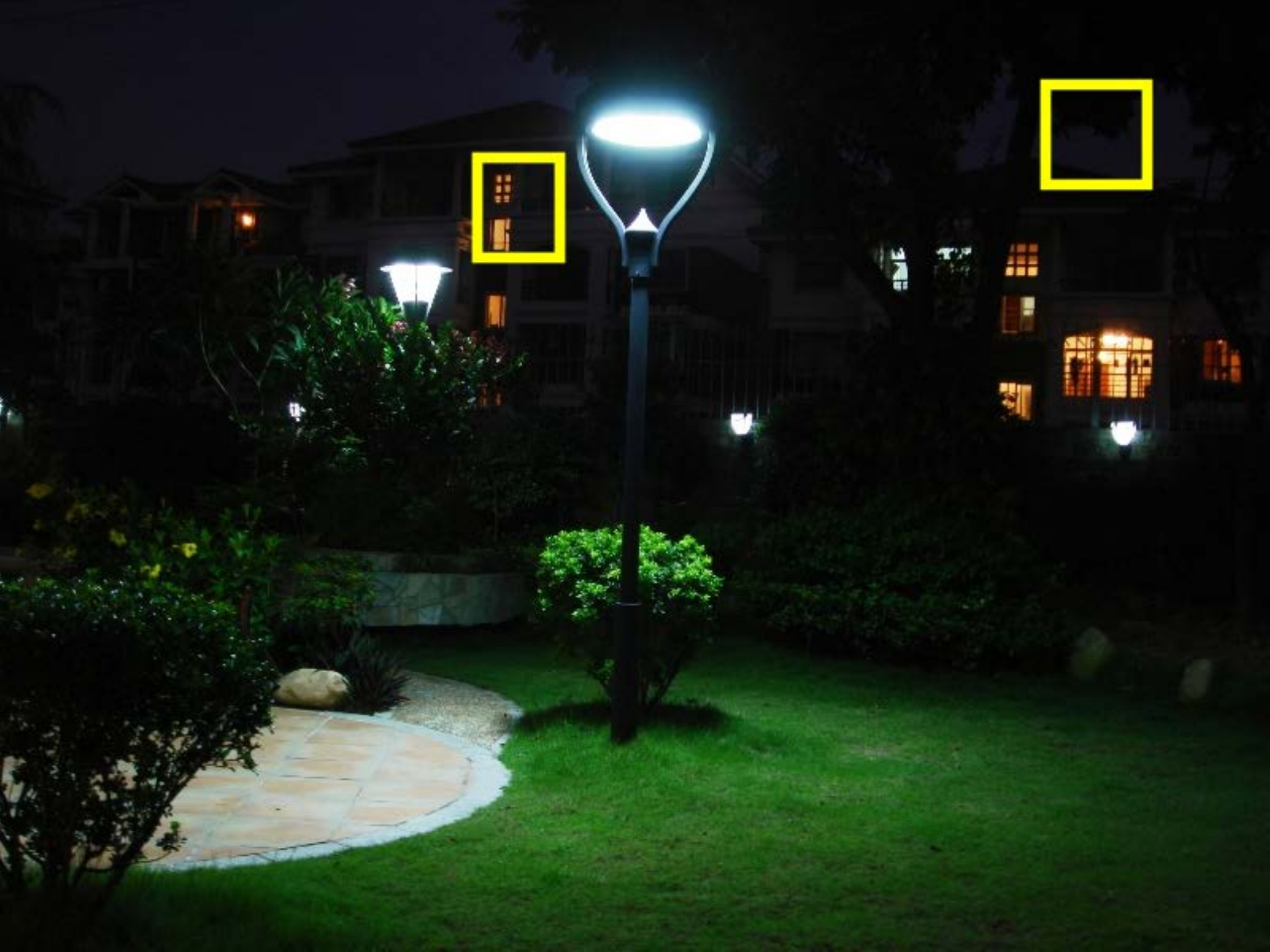} \vspace{11.9mm} \\
            \includegraphics[width=1\textwidth]{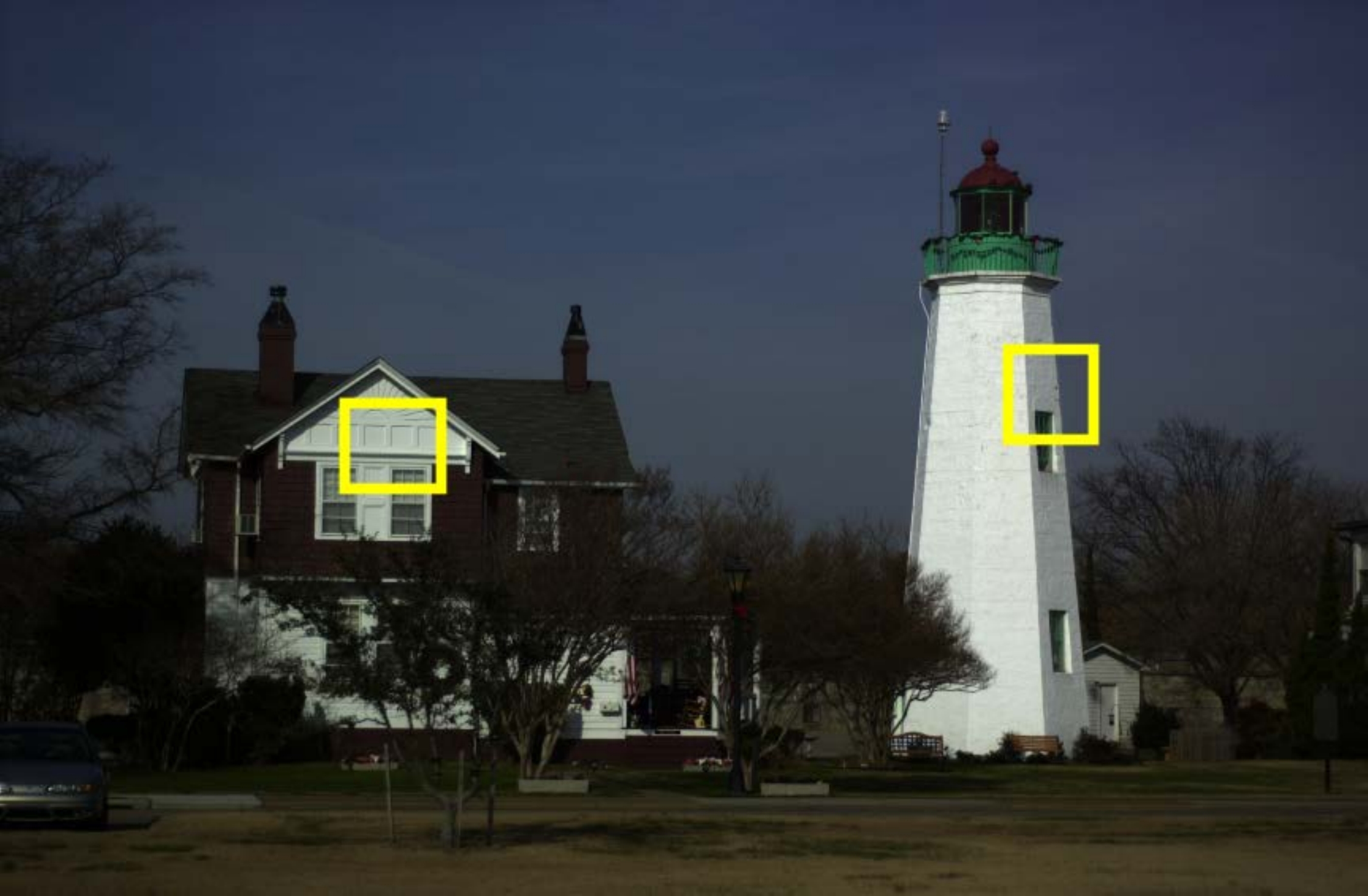} \vspace{8.9mm}
            \end{minipage}
        }\hspace{-3mm}
        \subfloat[]{
            \begin{minipage}[b]{0.1\textwidth}
                \includegraphics[width=1\textwidth]{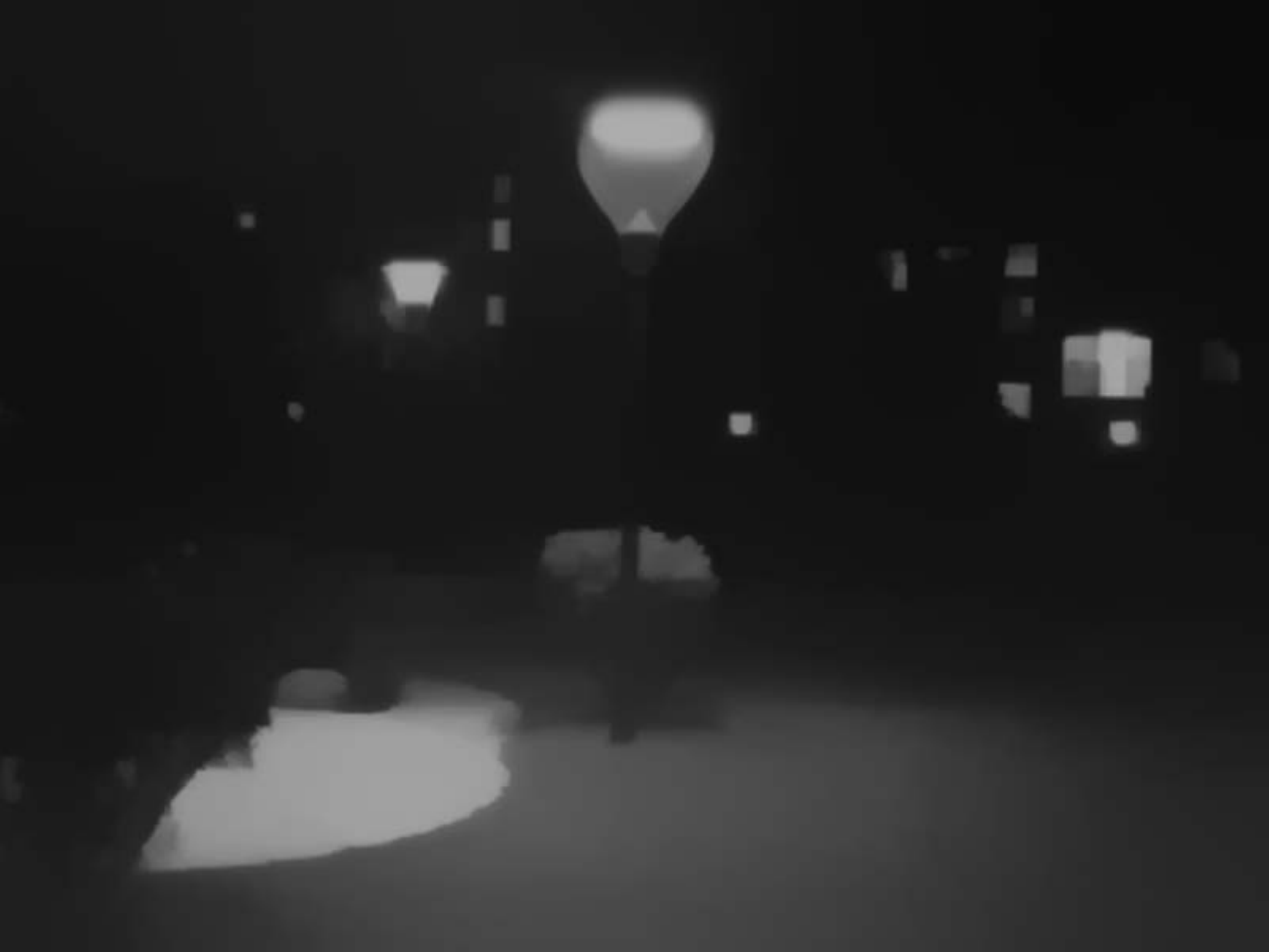} \vspace{-3mm} \\
                \includegraphics[width=1\textwidth]{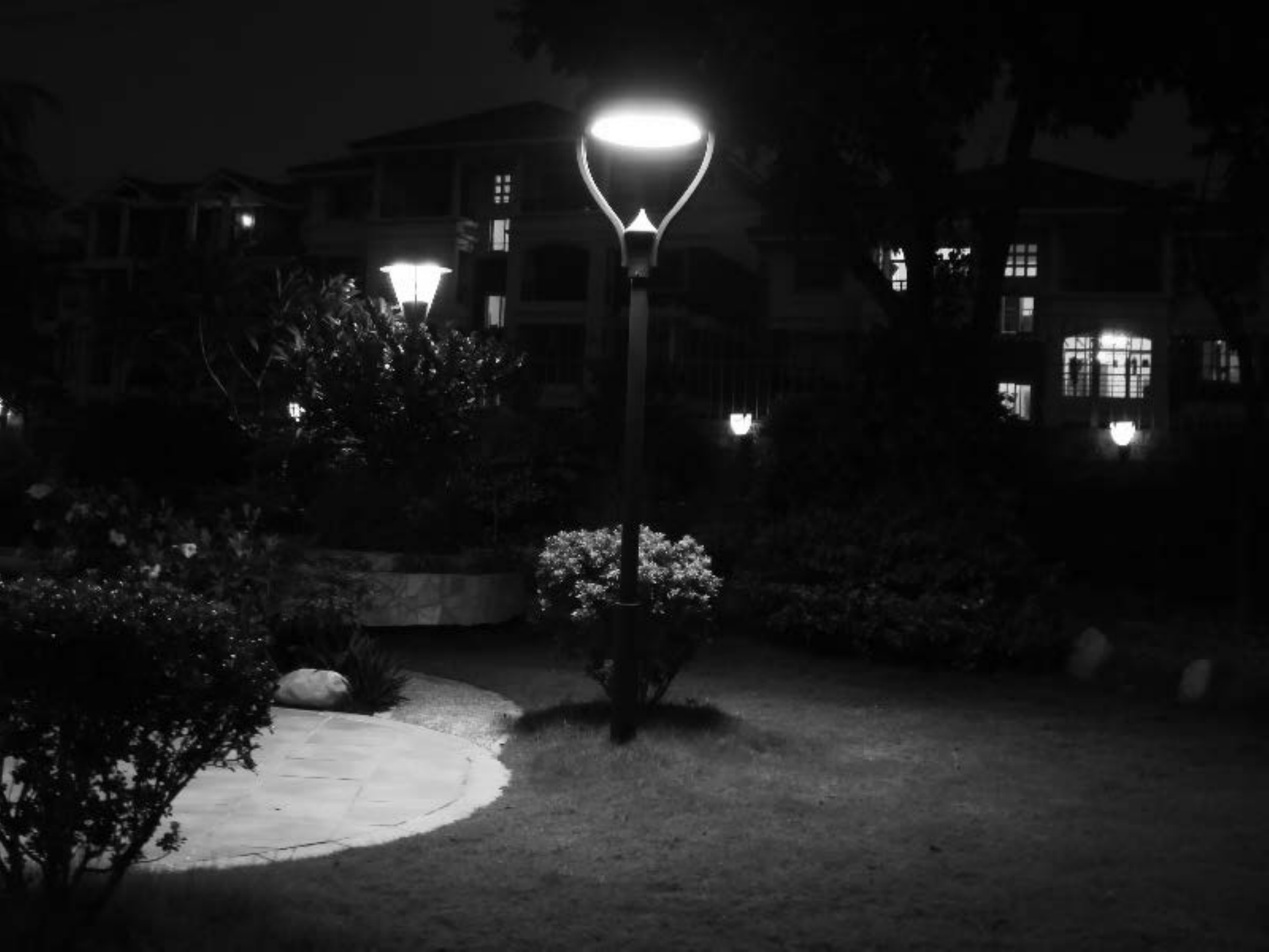} \vspace{-3mm} \\
                \includegraphics[width=1\textwidth]{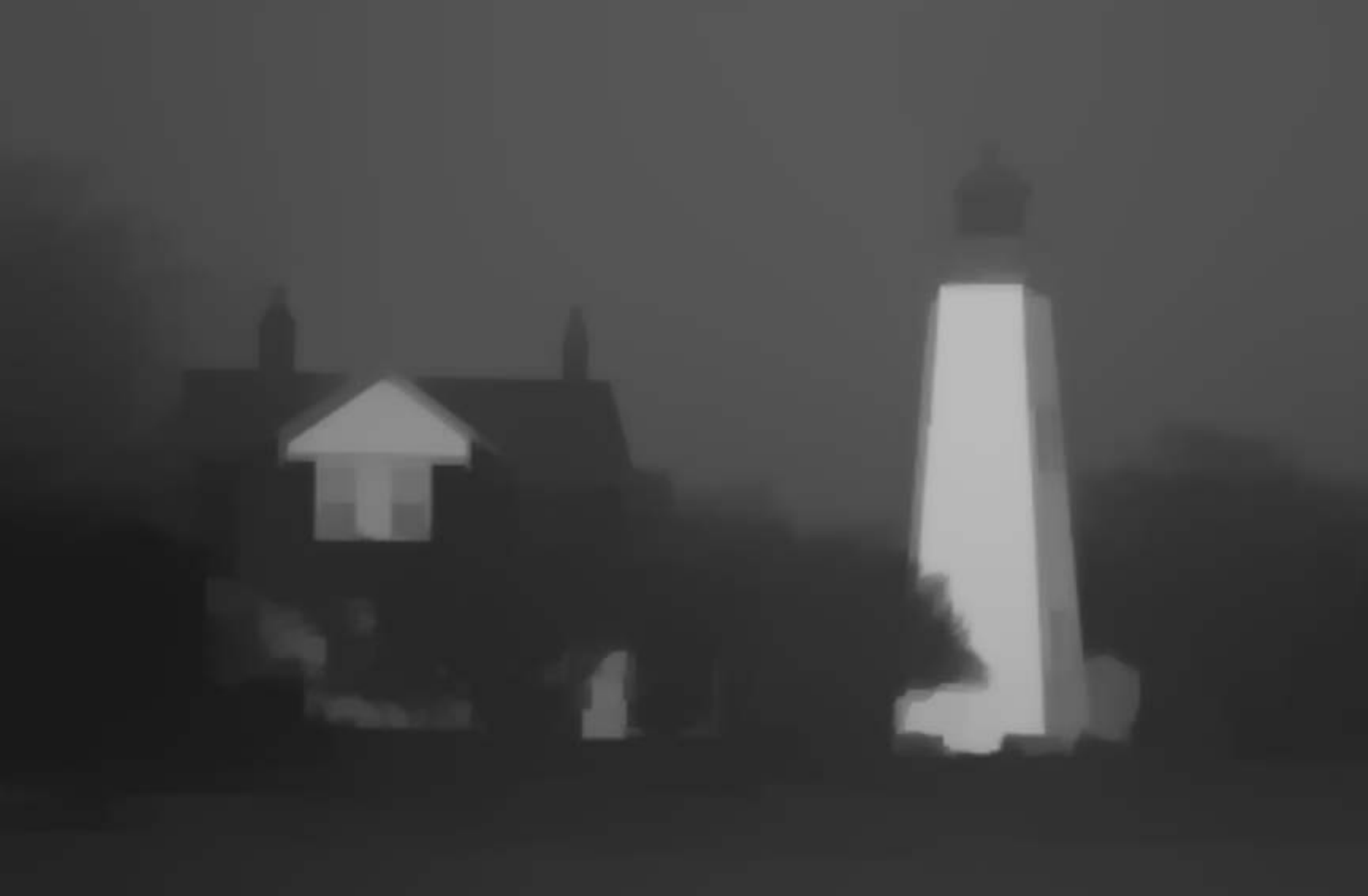} \vspace{-3mm} \\
                \includegraphics[width=1\textwidth]{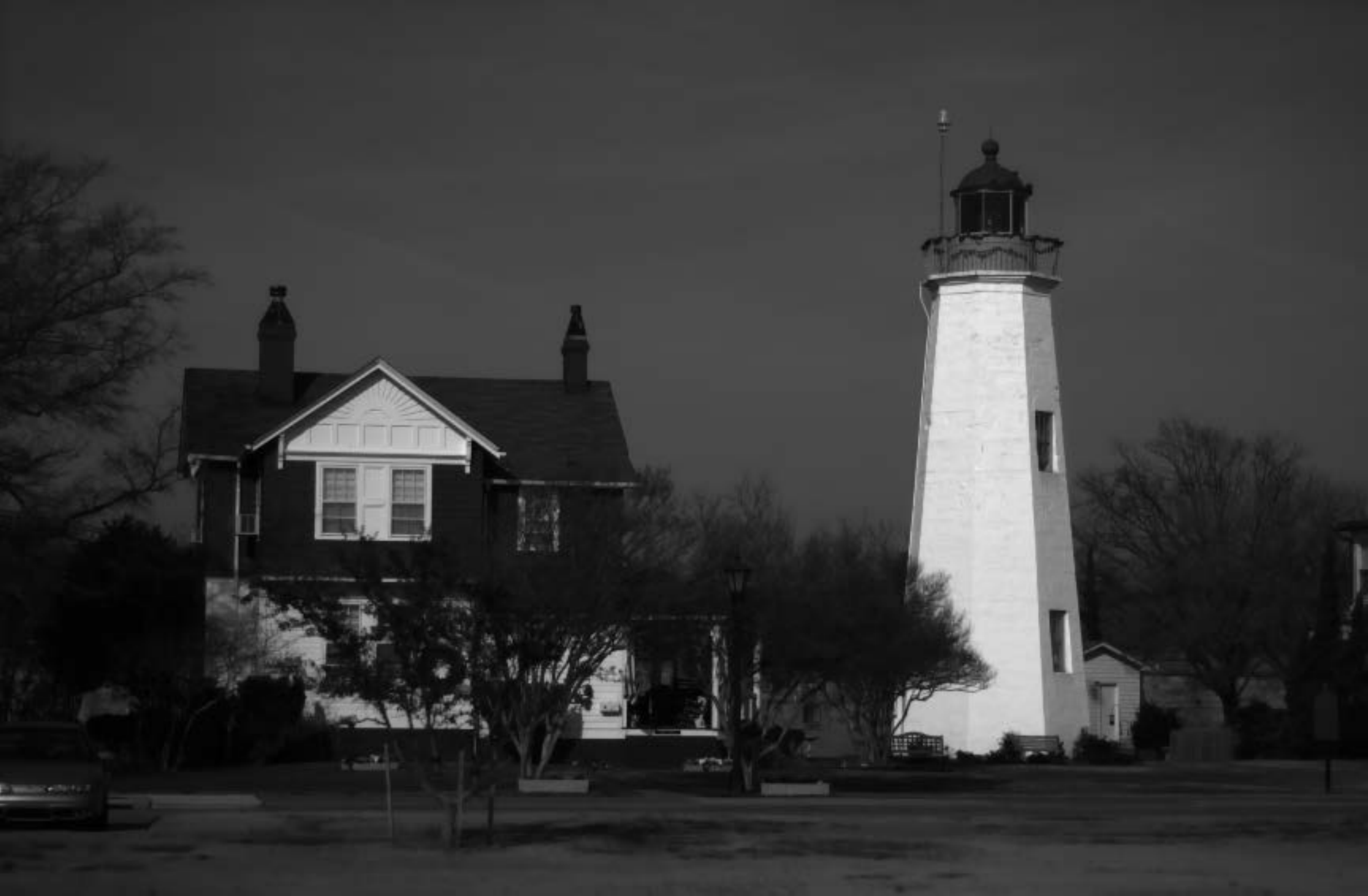}
            \end{minipage}
        }\hspace{-3mm}
        \subfloat[]{
            \begin{minipage}[b]{0.1\textwidth}
                \includegraphics[width=1\textwidth]{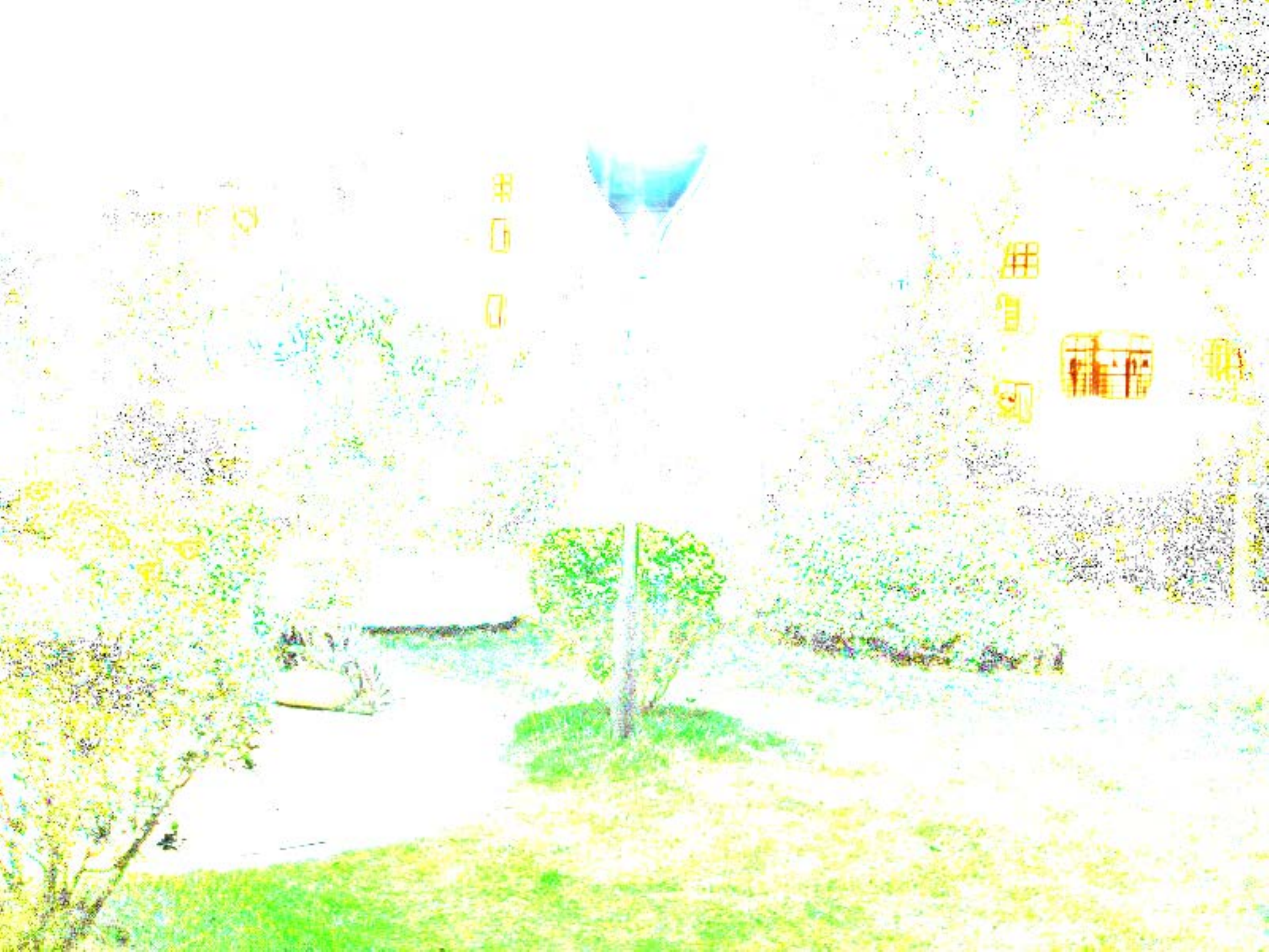} \vspace{-3mm} \\
                \includegraphics[width=1\textwidth]{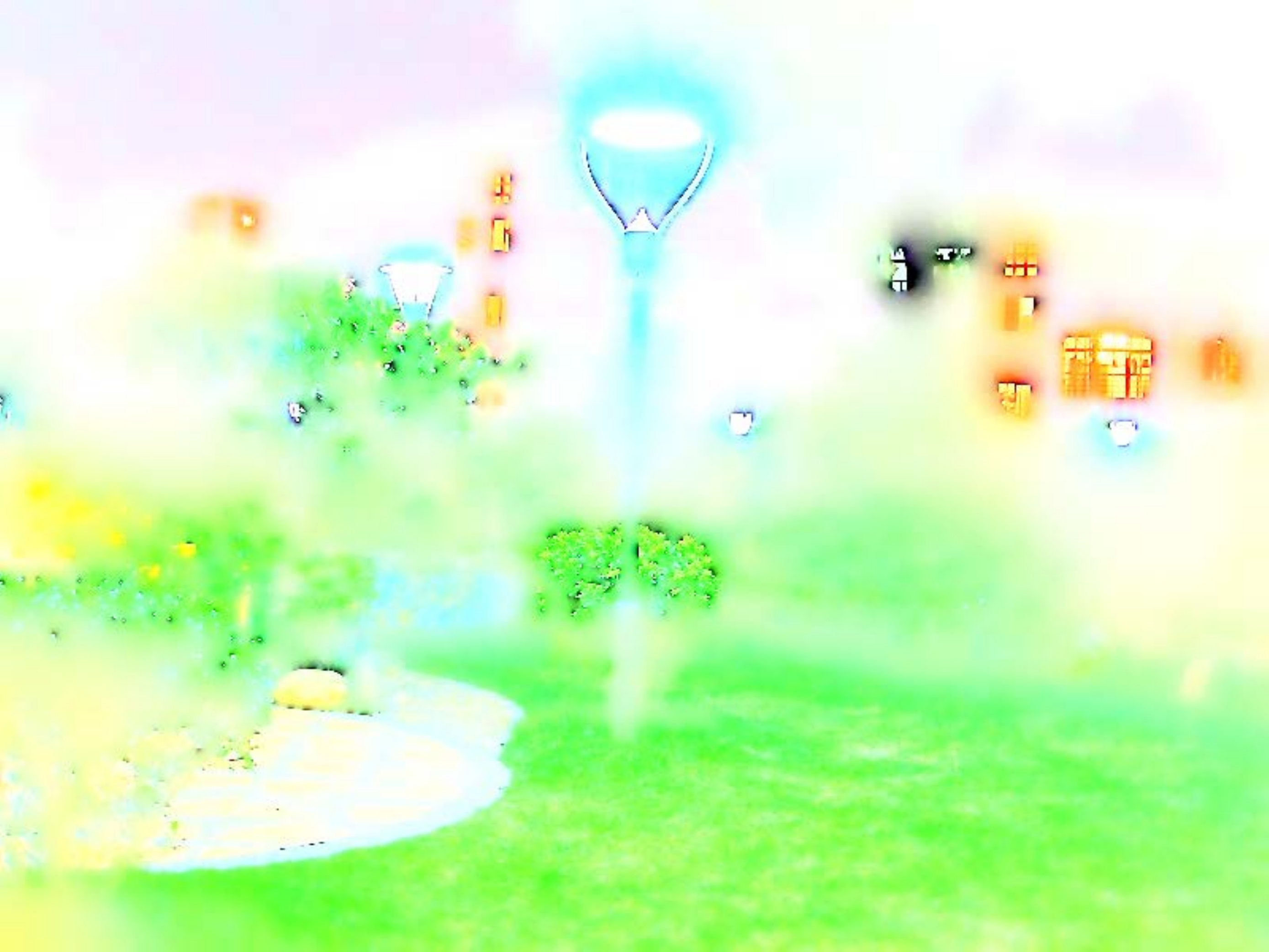} \vspace{-3mm} \\
                \includegraphics[width=1\textwidth]{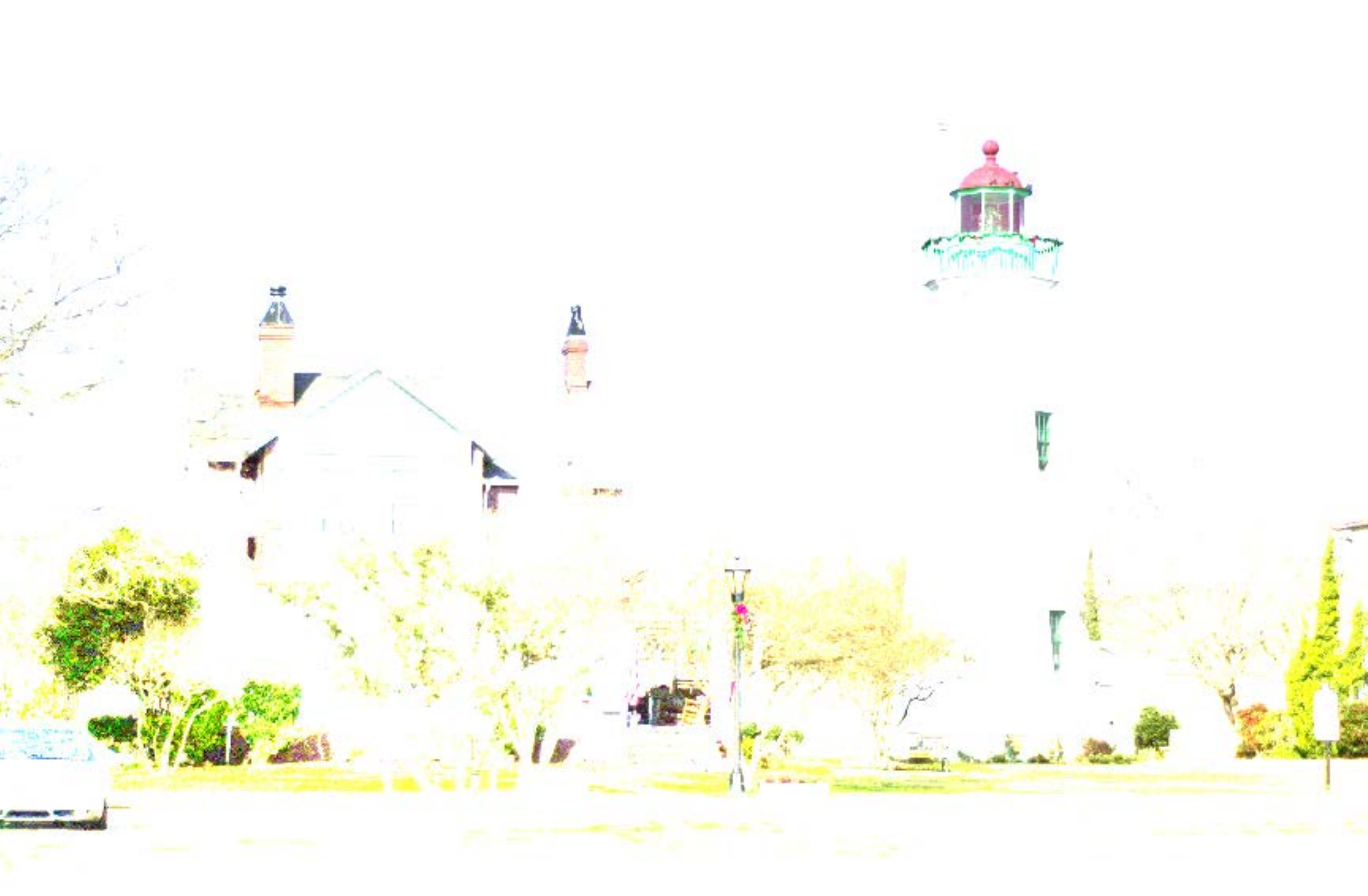} \vspace{-3mm} \\
                \includegraphics[width=1\textwidth]{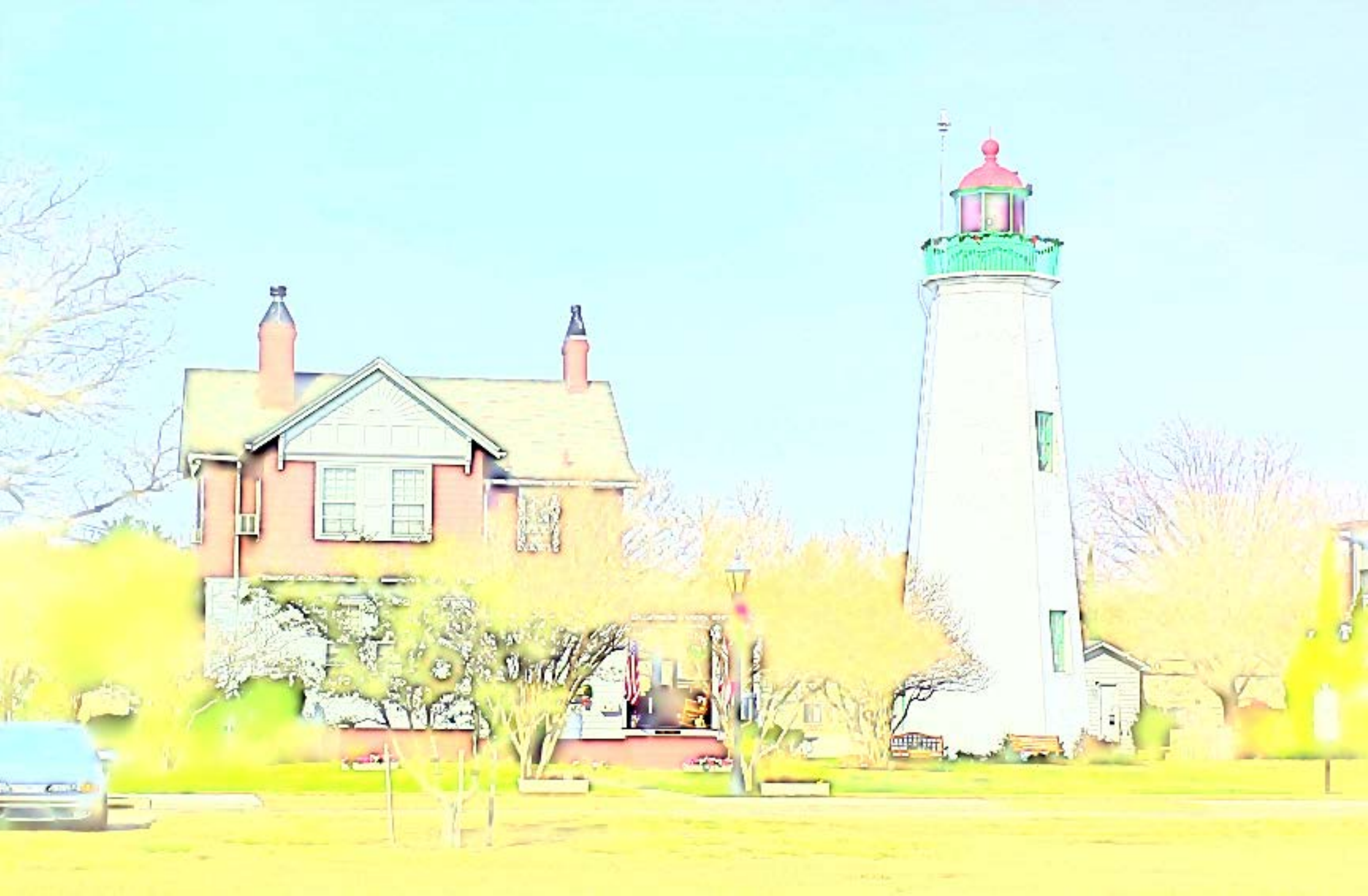}
            \end{minipage}
        }\hspace{-3mm}
        \subfloat[]{
            \begin{minipage}[b]{0.1\textwidth}
                \includegraphics[width=1\textwidth]{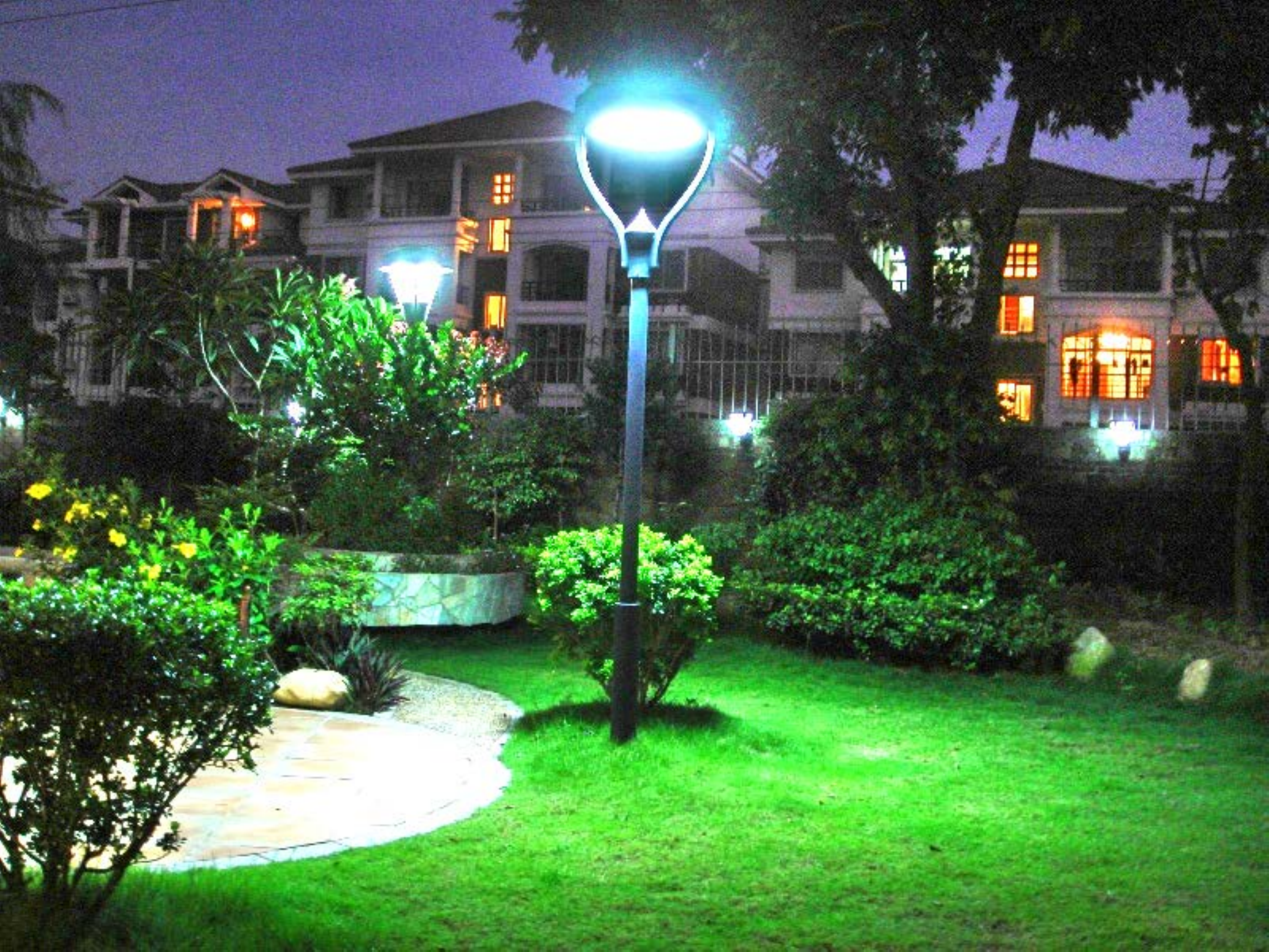} \vspace{-3mm} \\
                \includegraphics[width=1\textwidth]{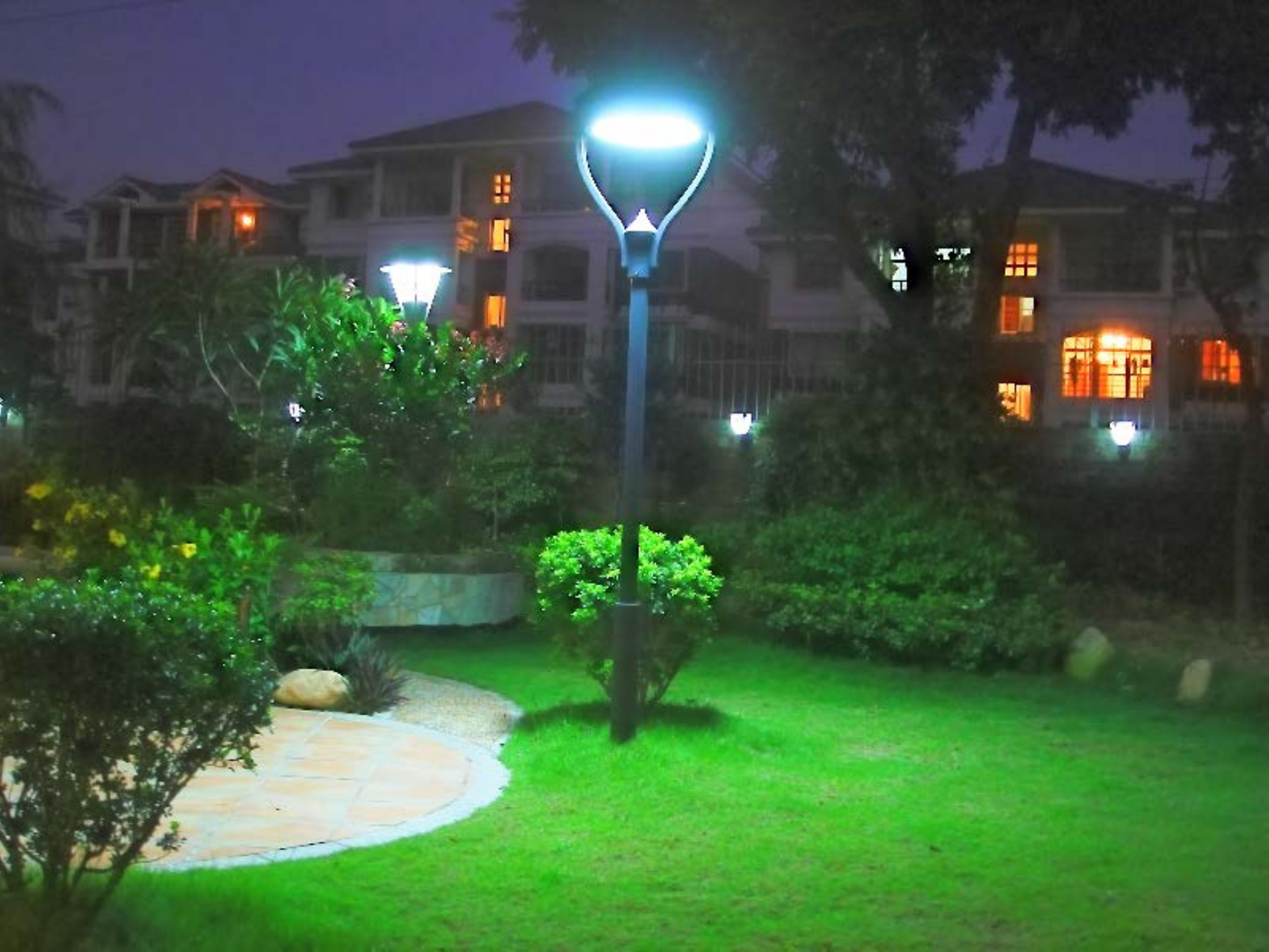} \vspace{-3mm} \\
                \includegraphics[width=1\textwidth]{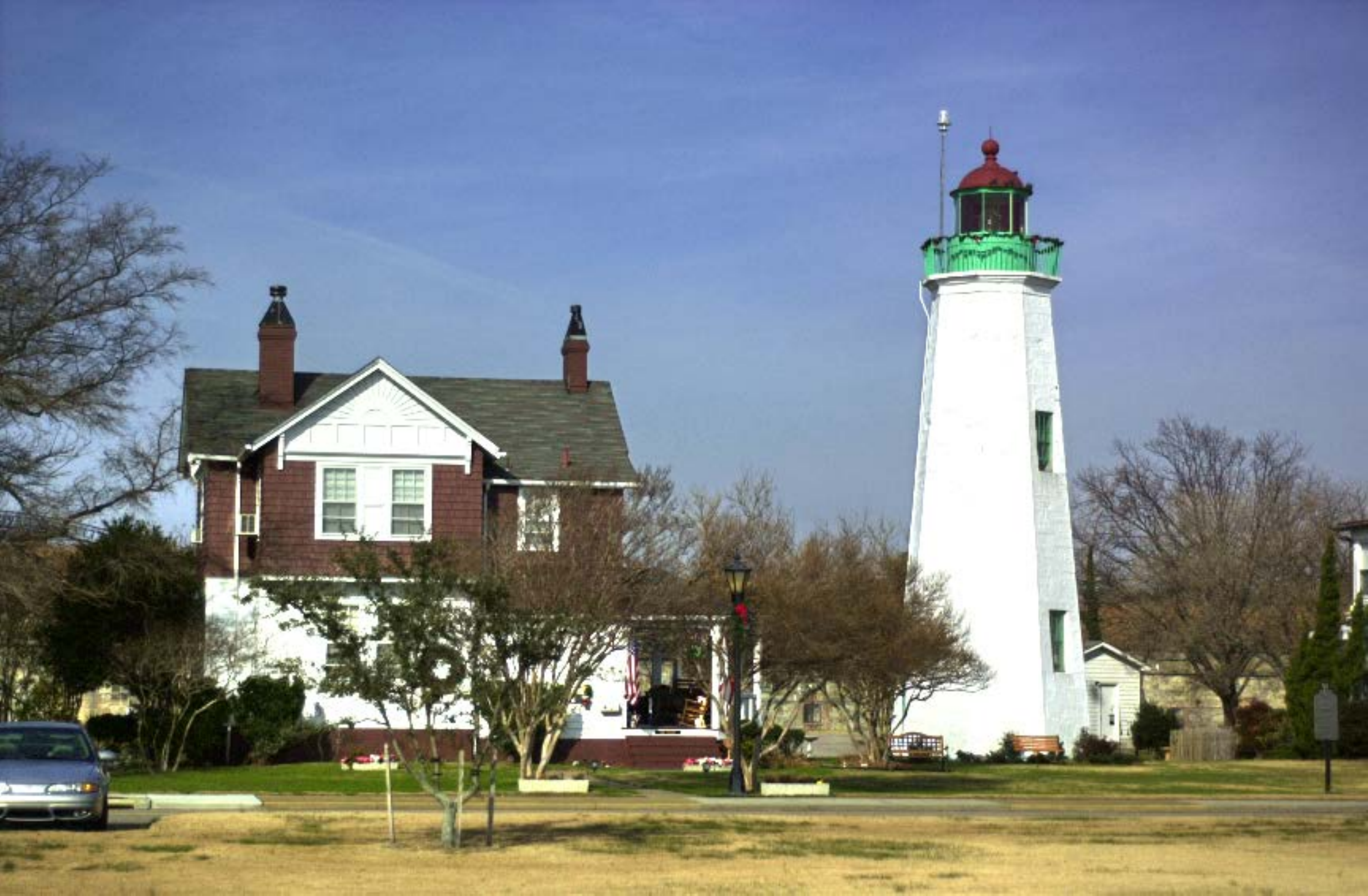} \vspace{-3mm} \\
                \includegraphics[width=1\textwidth]{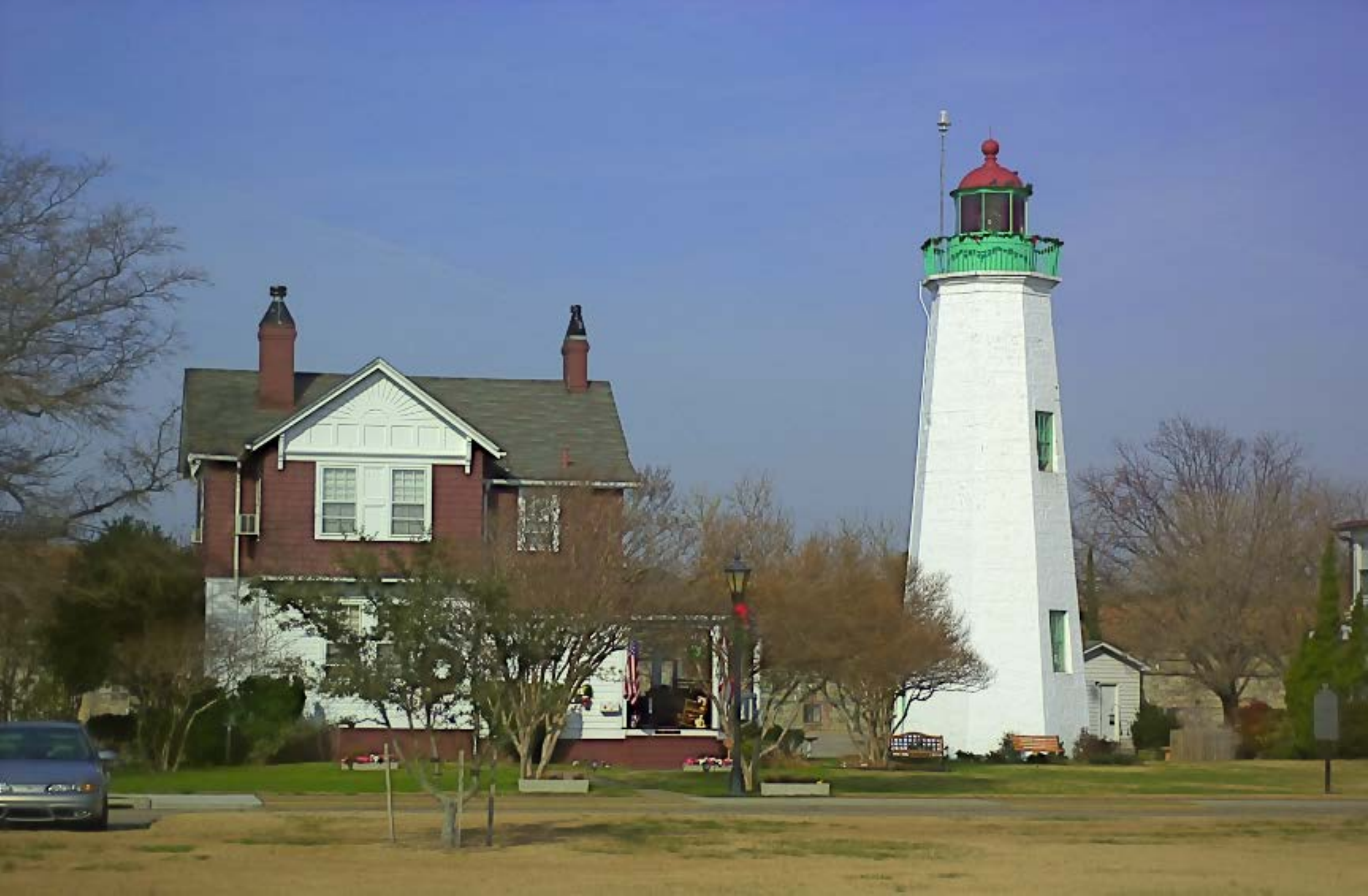}
            \end{minipage}
        }\hspace{-3mm}
        \subfloat[]{
            \begin{minipage}[b]{0.04\textwidth}
                \includegraphics[width=0.8\textwidth,height = 0.9\textwidth]{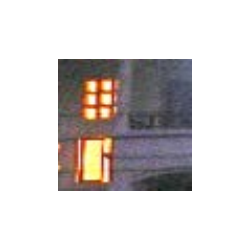} \vspace{0.5mm} \\
                \includegraphics[width=0.8\textwidth,height = 0.9\textwidth]{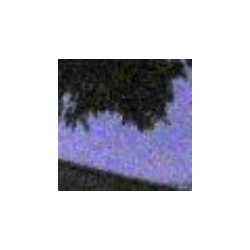} \vspace{1.3mm} \\
                \includegraphics[width=0.8\textwidth,height = 0.9\textwidth]{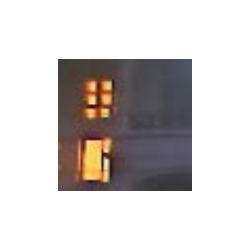} \vspace{0.5mm} \\
                \includegraphics[width=0.8\textwidth,height = 0.9\textwidth]{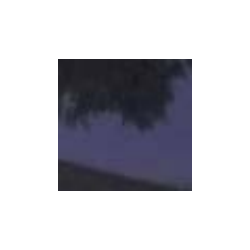} \vspace{1.2mm} \\
                \includegraphics[width=0.8\textwidth,height = 0.8\textwidth]{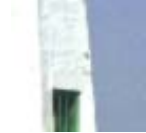} \vspace{0.3mm} \\
                \includegraphics[width=0.8\textwidth,height = 0.8\textwidth]{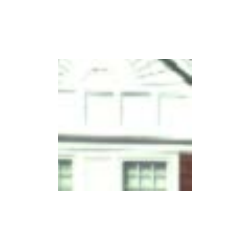} \vspace{1.2mm} \\
                \includegraphics[width=0.8\textwidth,height = 0.8\textwidth]{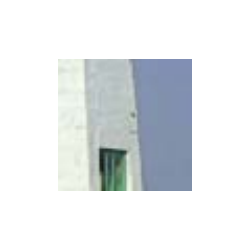} \vspace{0.3mm} \\
                \includegraphics[width=0.8\textwidth, height = 0.8\textwidth]{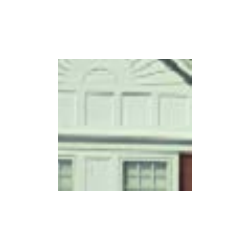}
            \end{minipage}
        }
        \caption{Comparison of the illumination, reflectance and result images of our method and LIME \cite{7782813} with details. (a),(b),(c),(d) are the input image, the illumination, reflectance, and the result images, respectively. (e) is the details of the result images. In each case from top to bottom: results of LIME\cite{7782813} and ours.}
        \label{F_decmp}
    \end{figure}

    \begin{figure*} [ht]
        \centering
        \subfloat[Original]{
            \begin{minipage}[b]{0.15\textwidth}
                \includegraphics[width=1\textwidth]{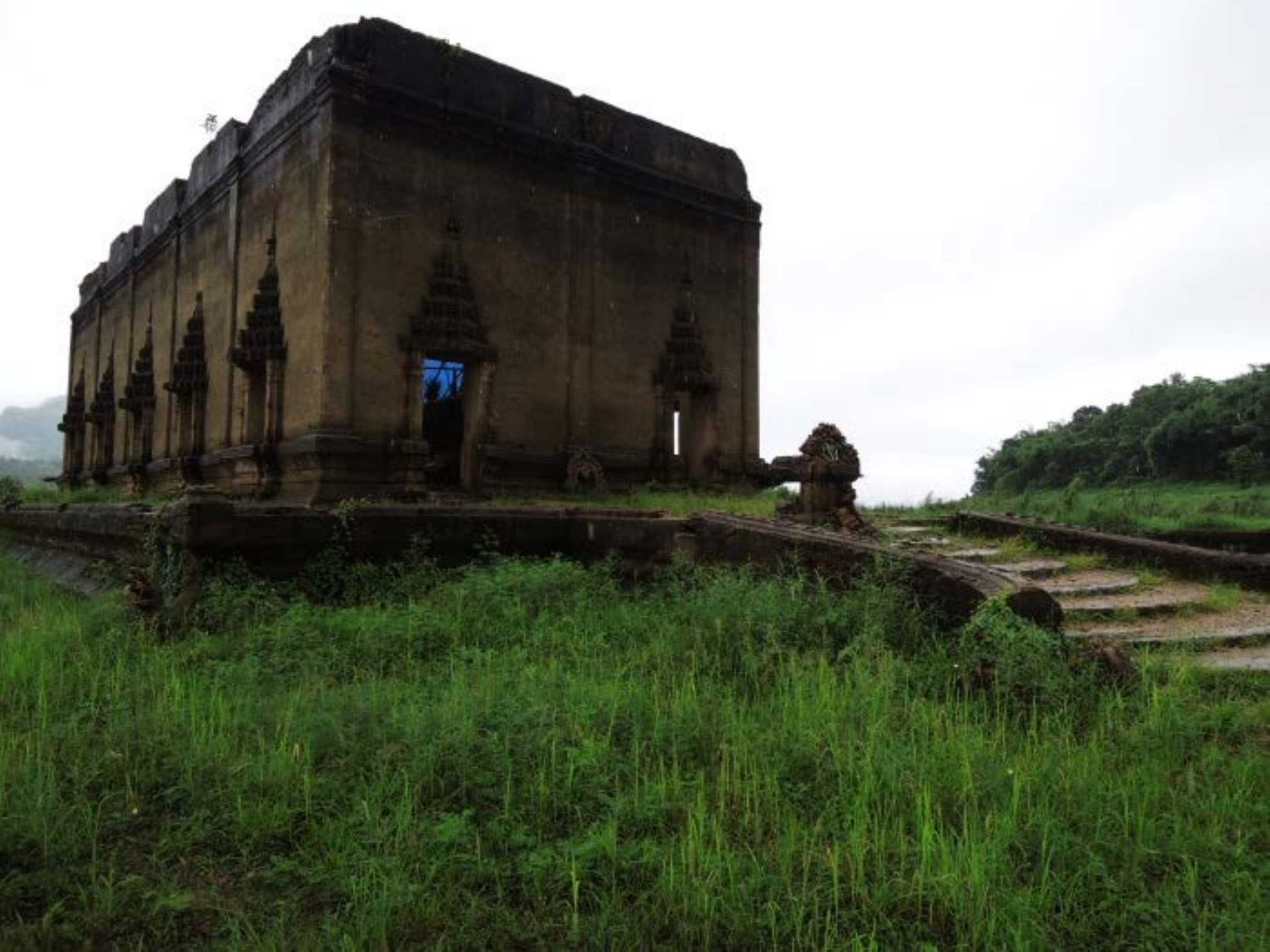} \vspace{-3mm} \\
                \includegraphics[width=1\textwidth]{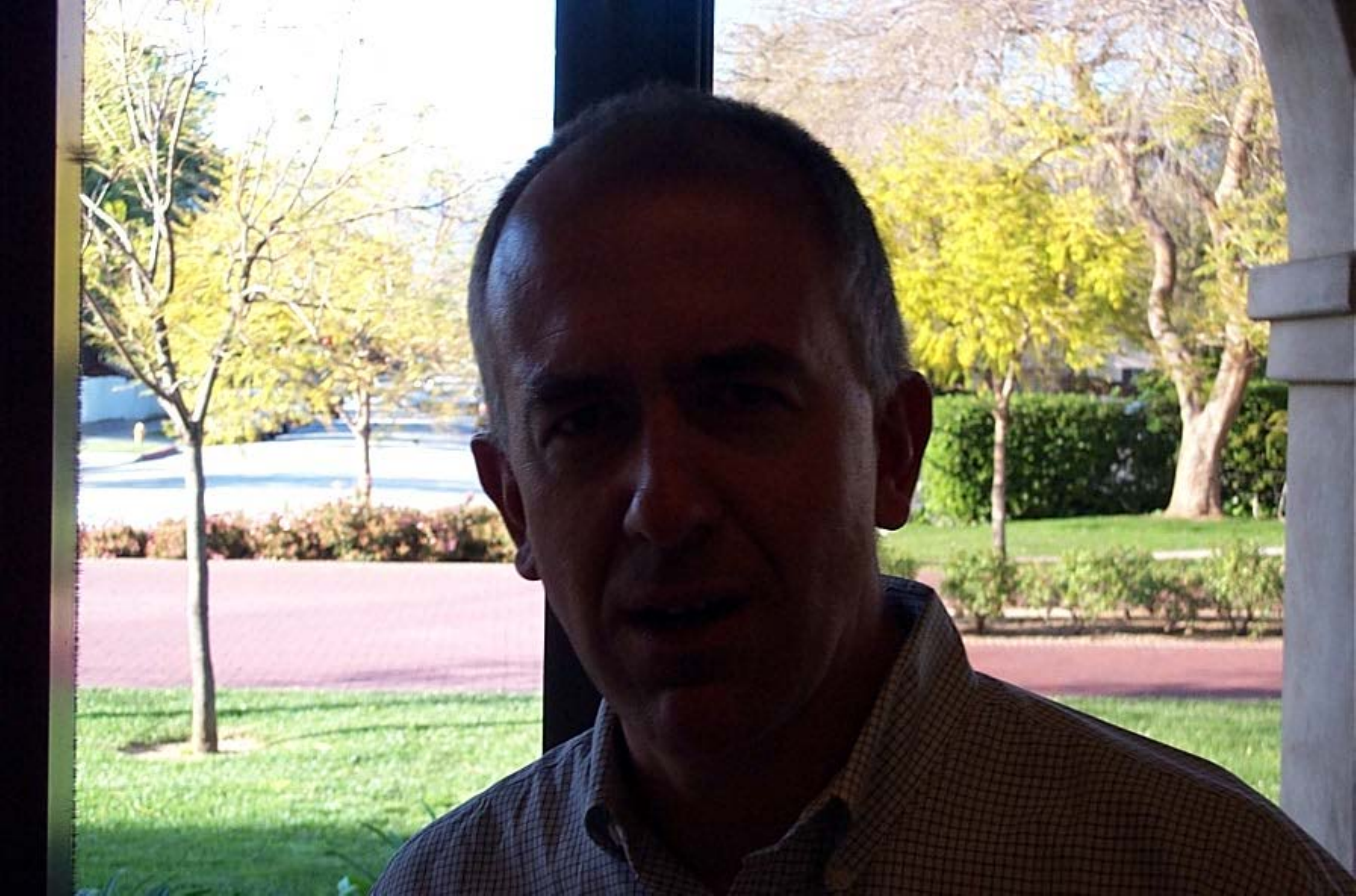} 
            \end{minipage}
        } \hspace{-3mm}
        \subfloat[HE]{
            \begin{minipage}[b]{0.15\textwidth}
                \includegraphics[width=1\textwidth]{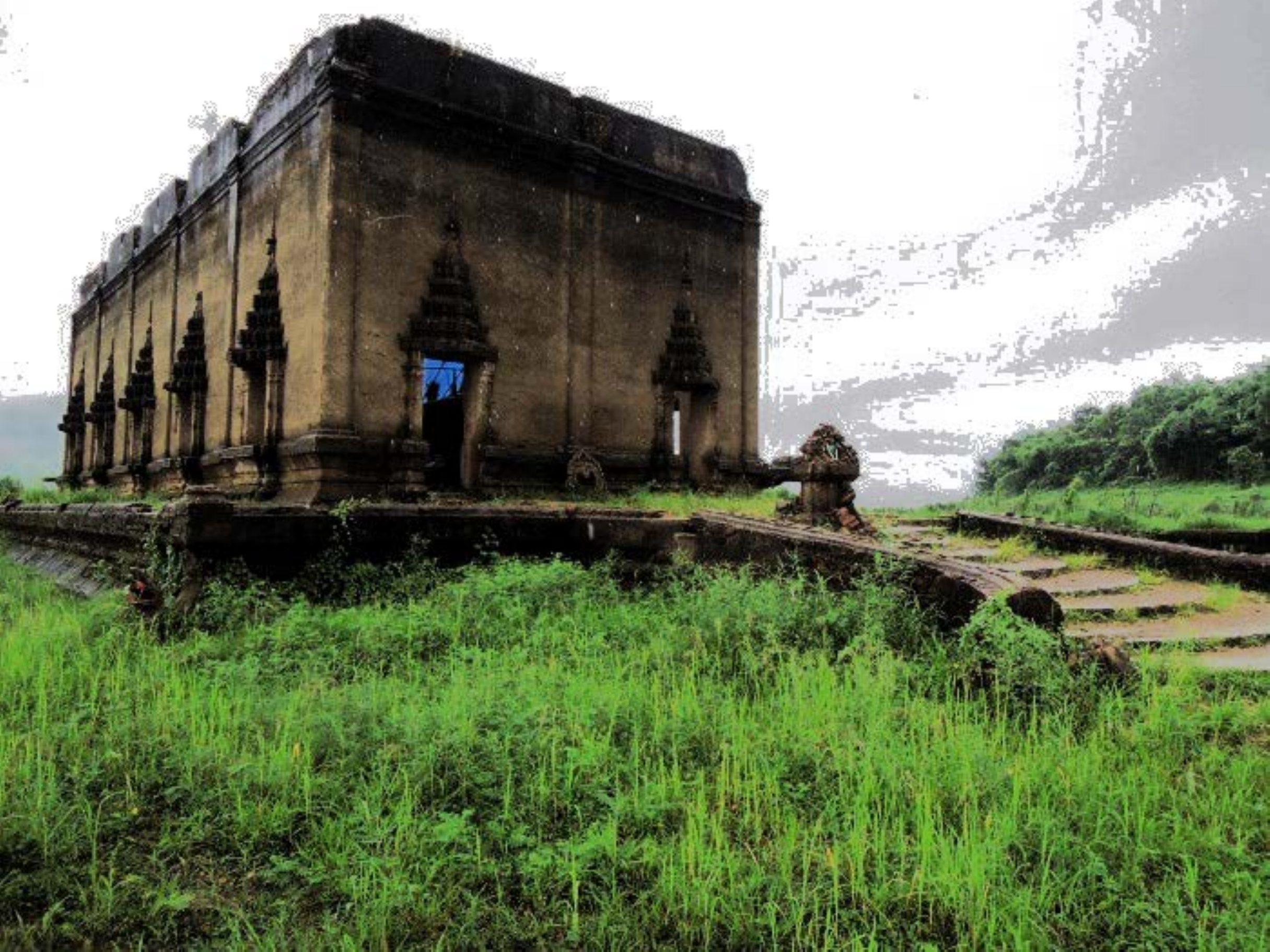} \vspace{-3mm} \\
                \includegraphics[width=1\textwidth]{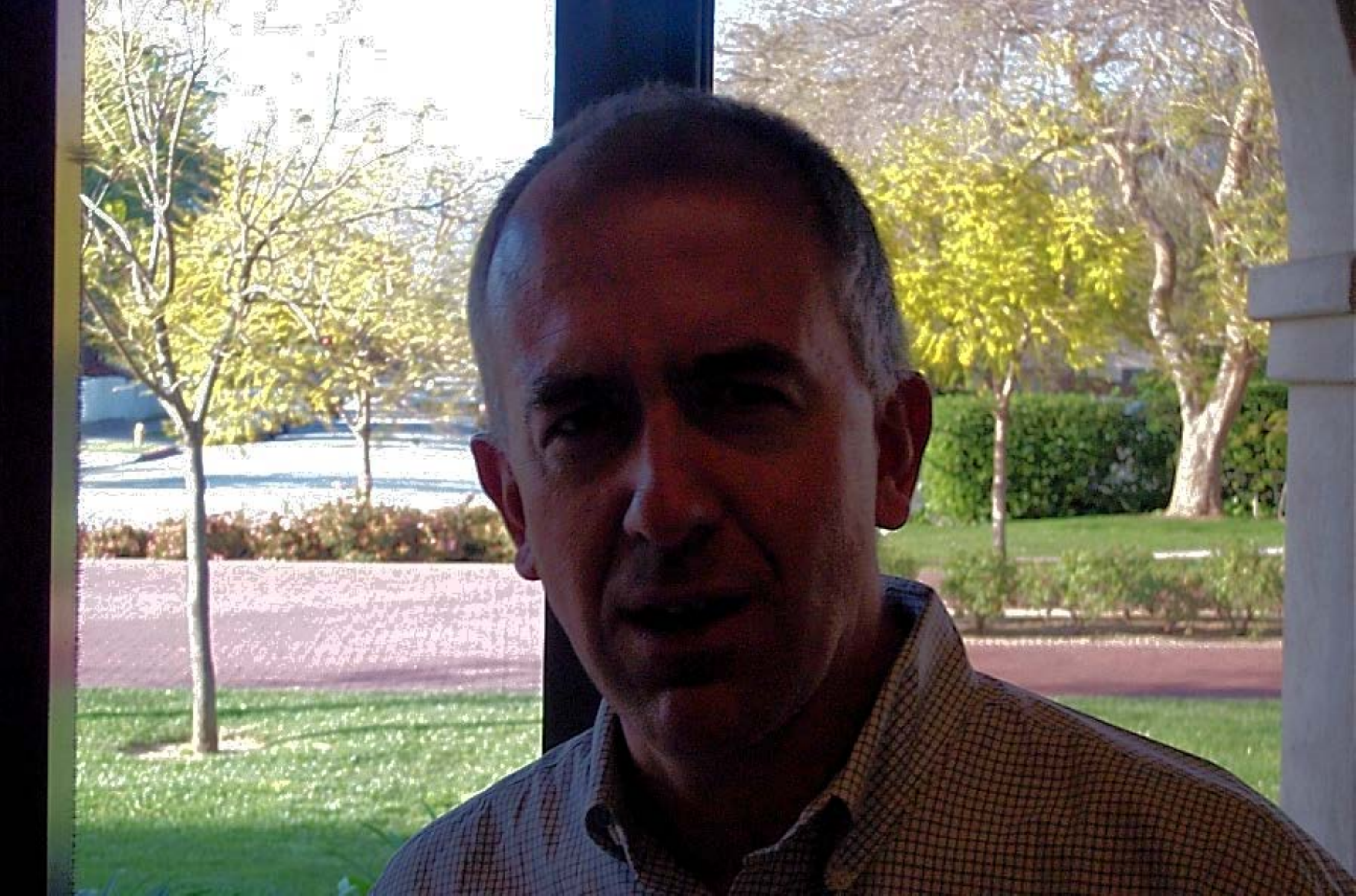} 
            \end{minipage}
        } \hspace{-3mm}
        \subfloat[SRIE \cite{7780673}]{
            \begin{minipage}[b]{0.15\textwidth}
                \includegraphics[width=1\textwidth]{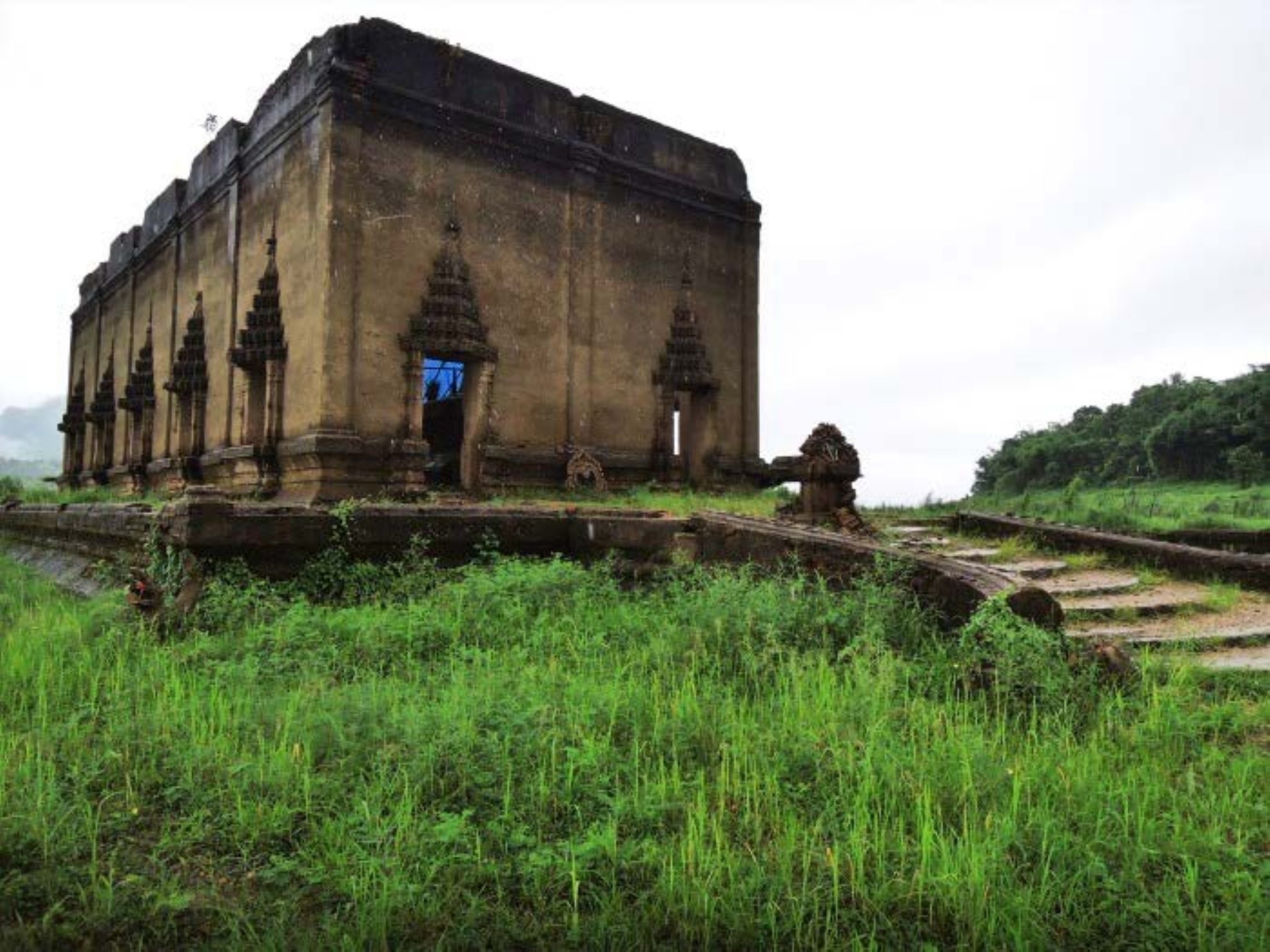} \vspace{-3mm} \\
                \includegraphics[width=1\textwidth]{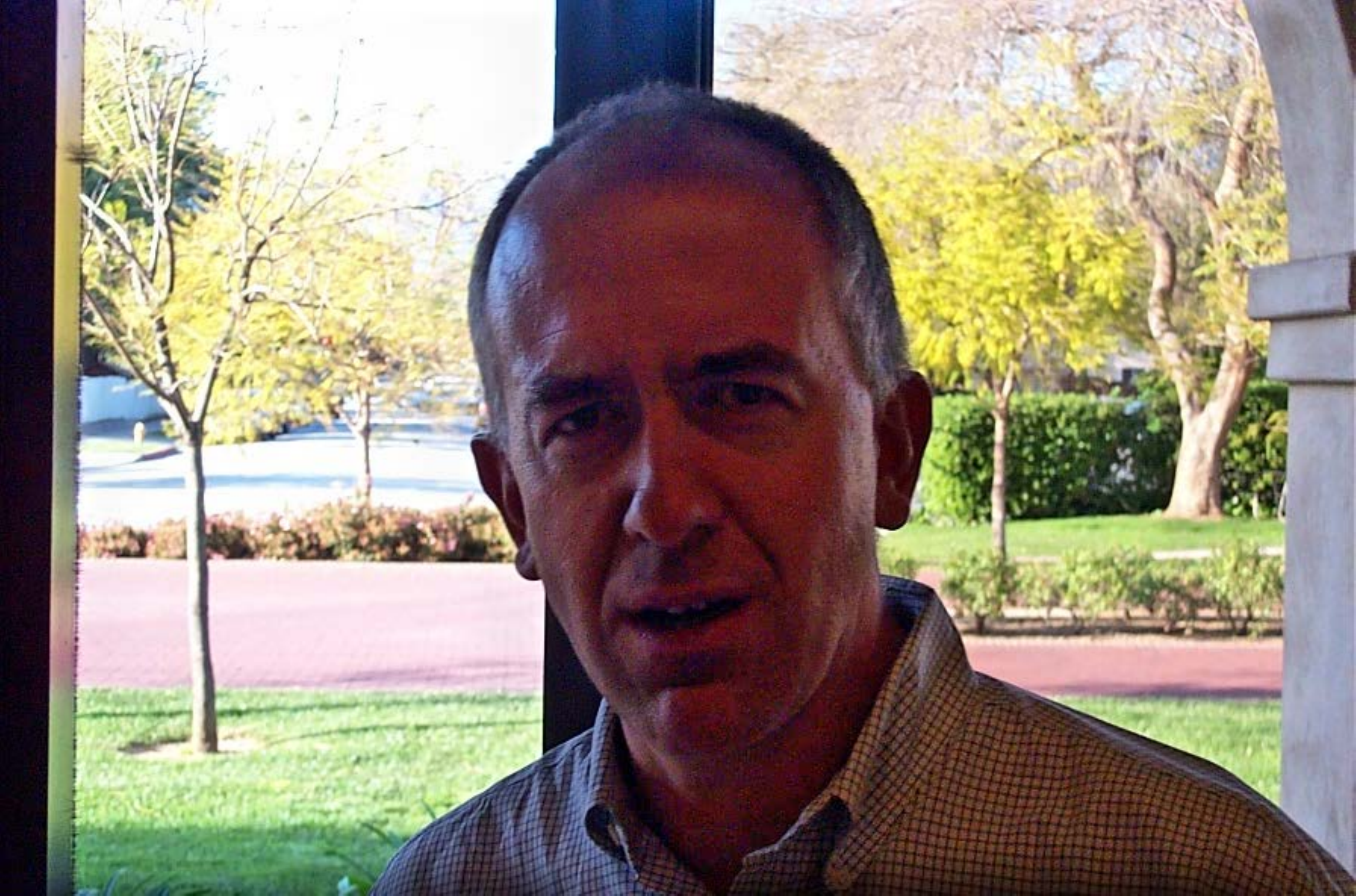} 
            \end{minipage}
        } \hspace{-3mm}
        \subfloat[NPEA \cite{6512558}]{
            \begin{minipage}[b]{0.15\textwidth}
                \includegraphics[width=1\textwidth]{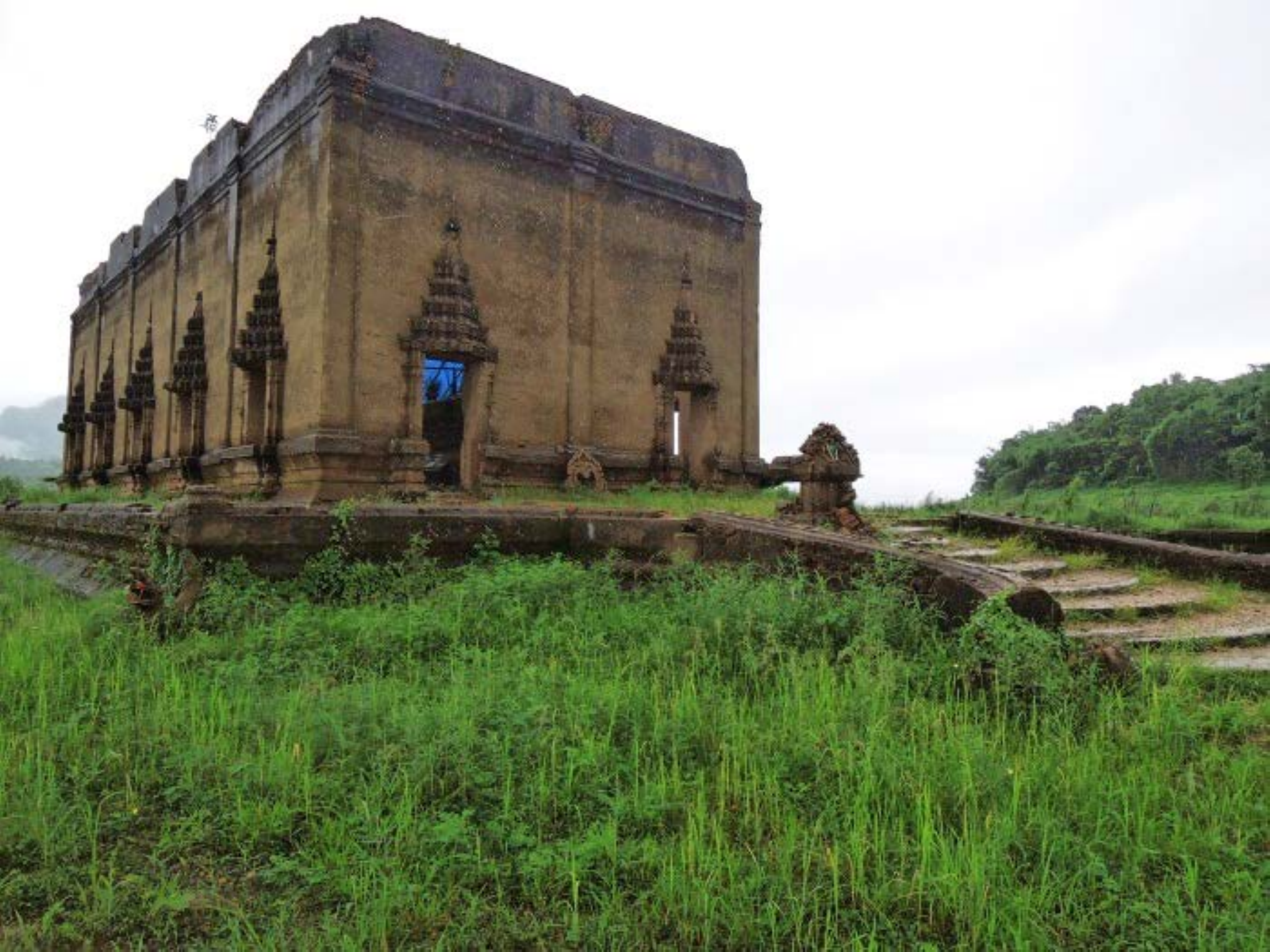} \vspace{-3mm} \\
                \includegraphics[width=1\textwidth]{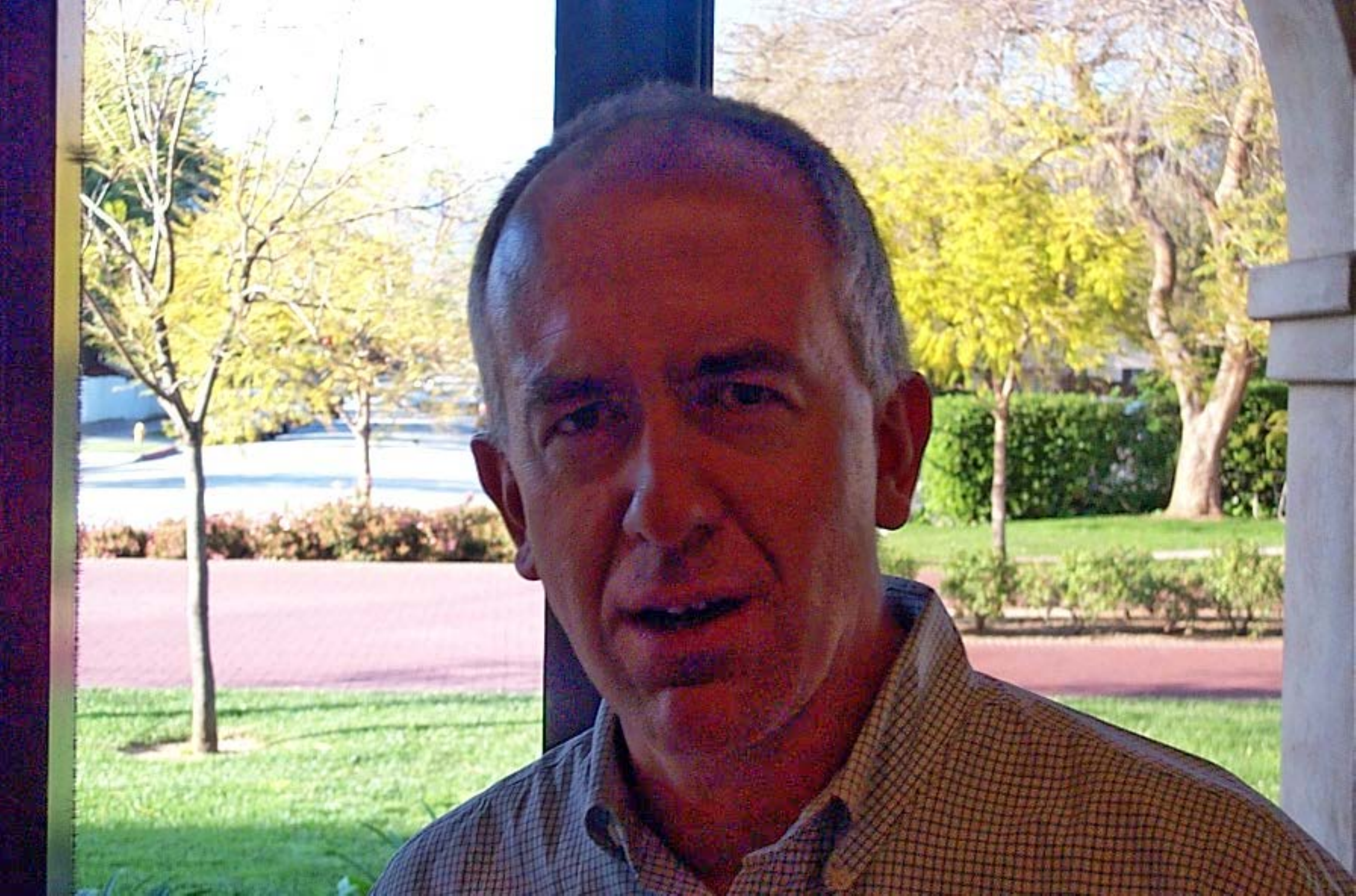} 
            \end{minipage}
        } \hspace{-3mm}
        \subfloat[LIME \cite{7782813}]{
            \begin{minipage}[b]{0.15\textwidth}
                \includegraphics[width=1\textwidth]{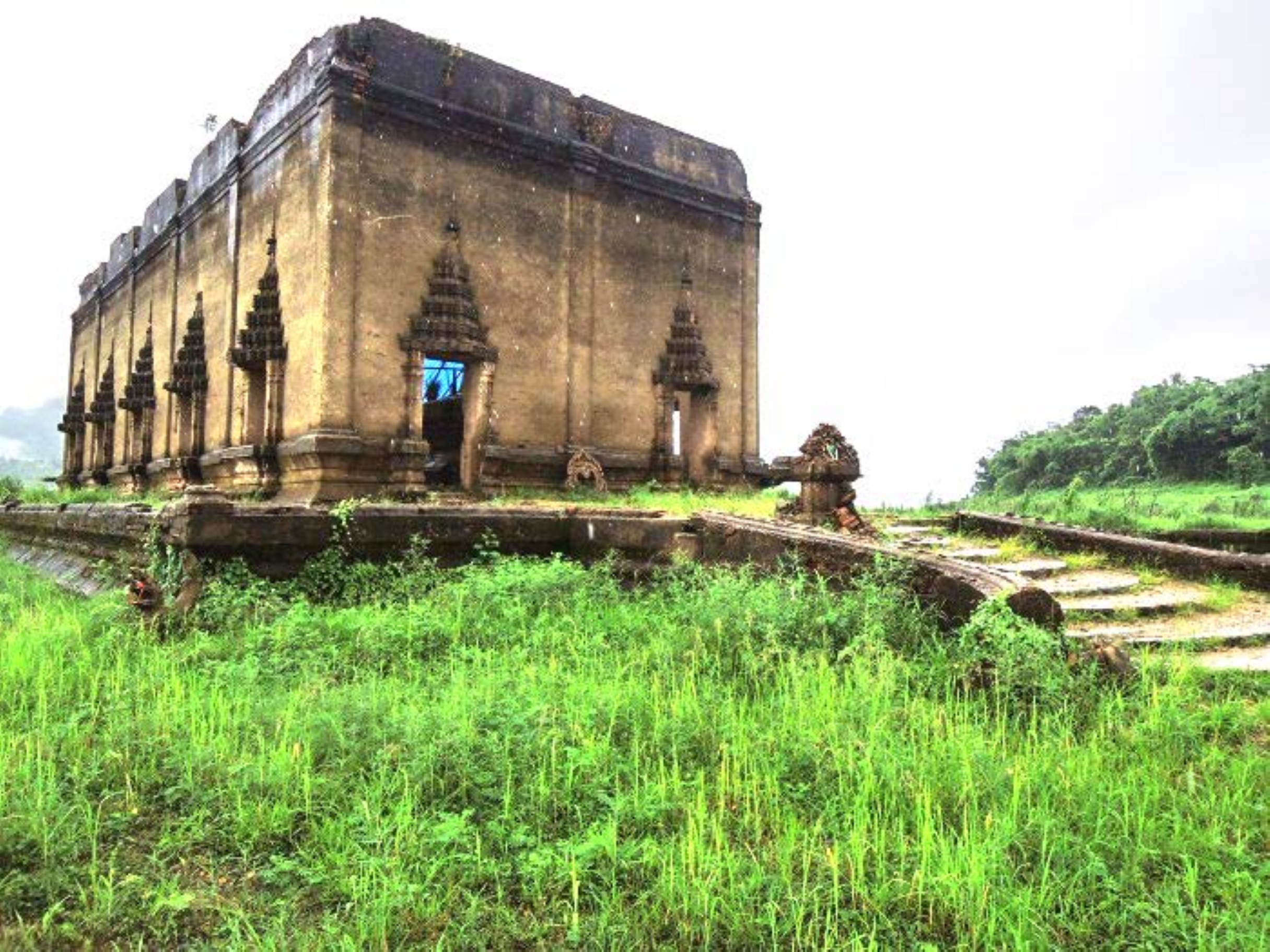} \vspace{-3mm} \\
                \includegraphics[width=1\textwidth]{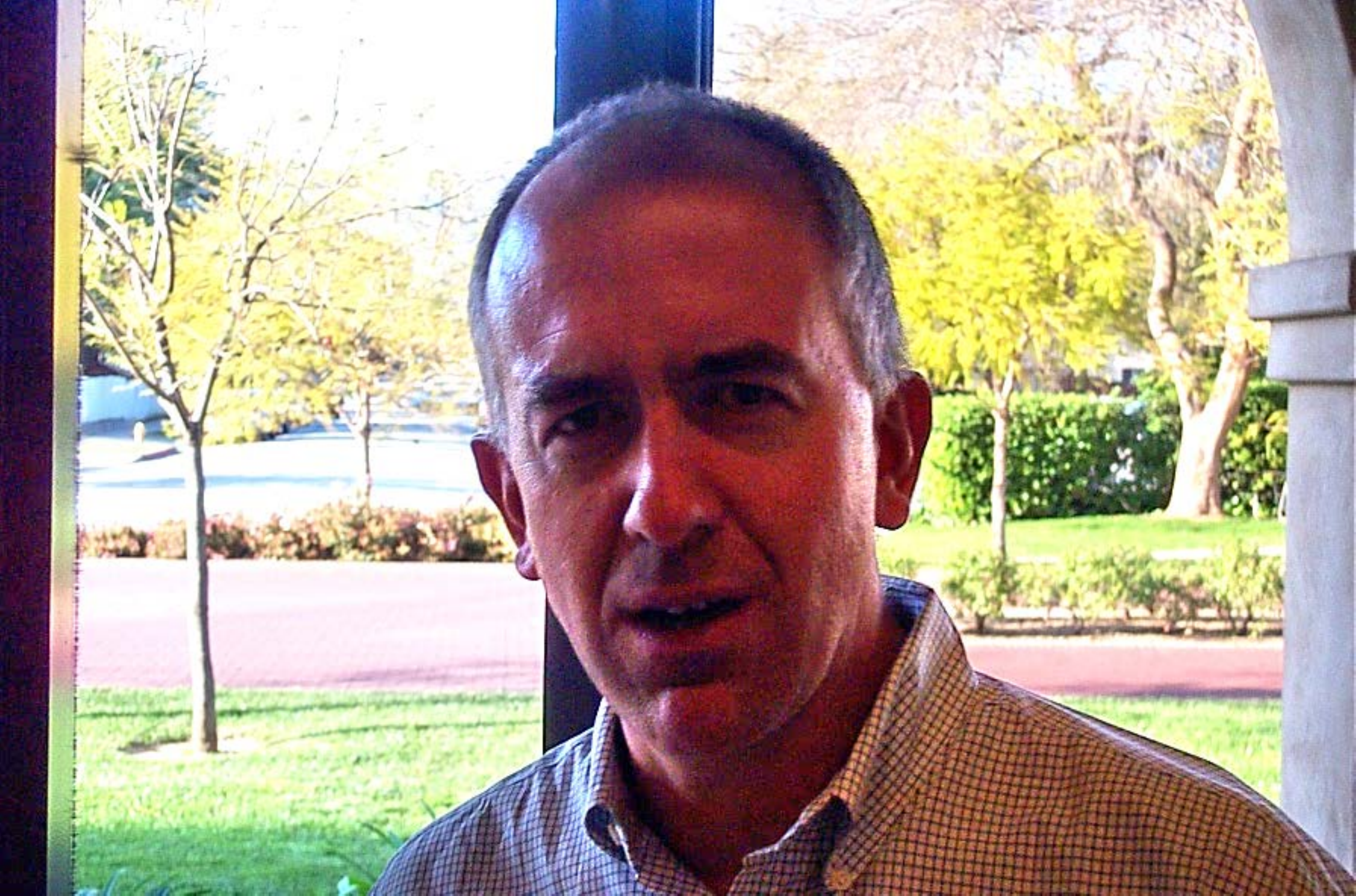} 
            \end{minipage}
        } \hspace{-3mm}
        \subfloat[Our Method]{
            \begin{minipage}[b]{0.15\textwidth}
                \includegraphics[width=1\textwidth]{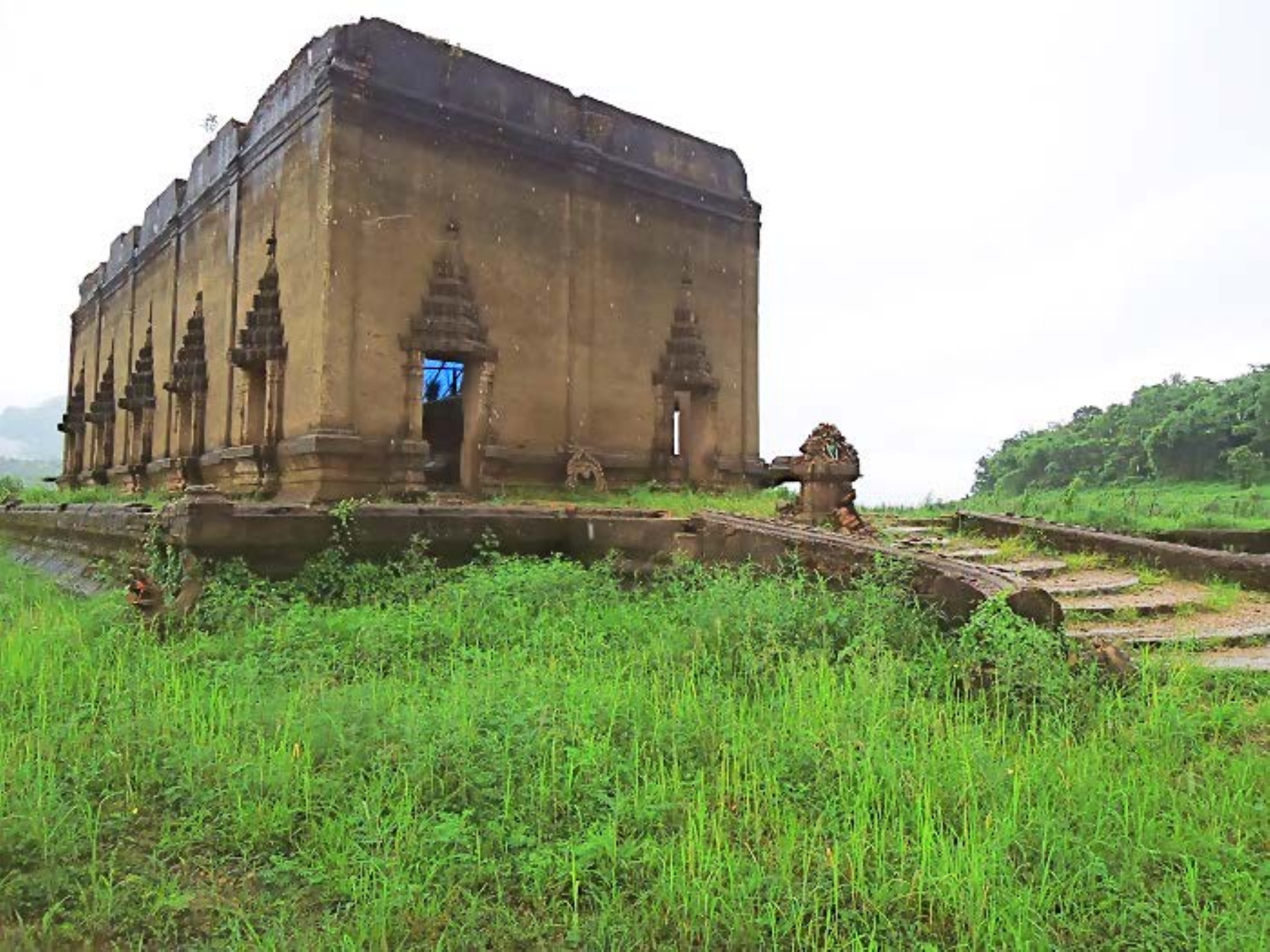} \vspace{-3mm} \\
                \includegraphics[width=1\textwidth]{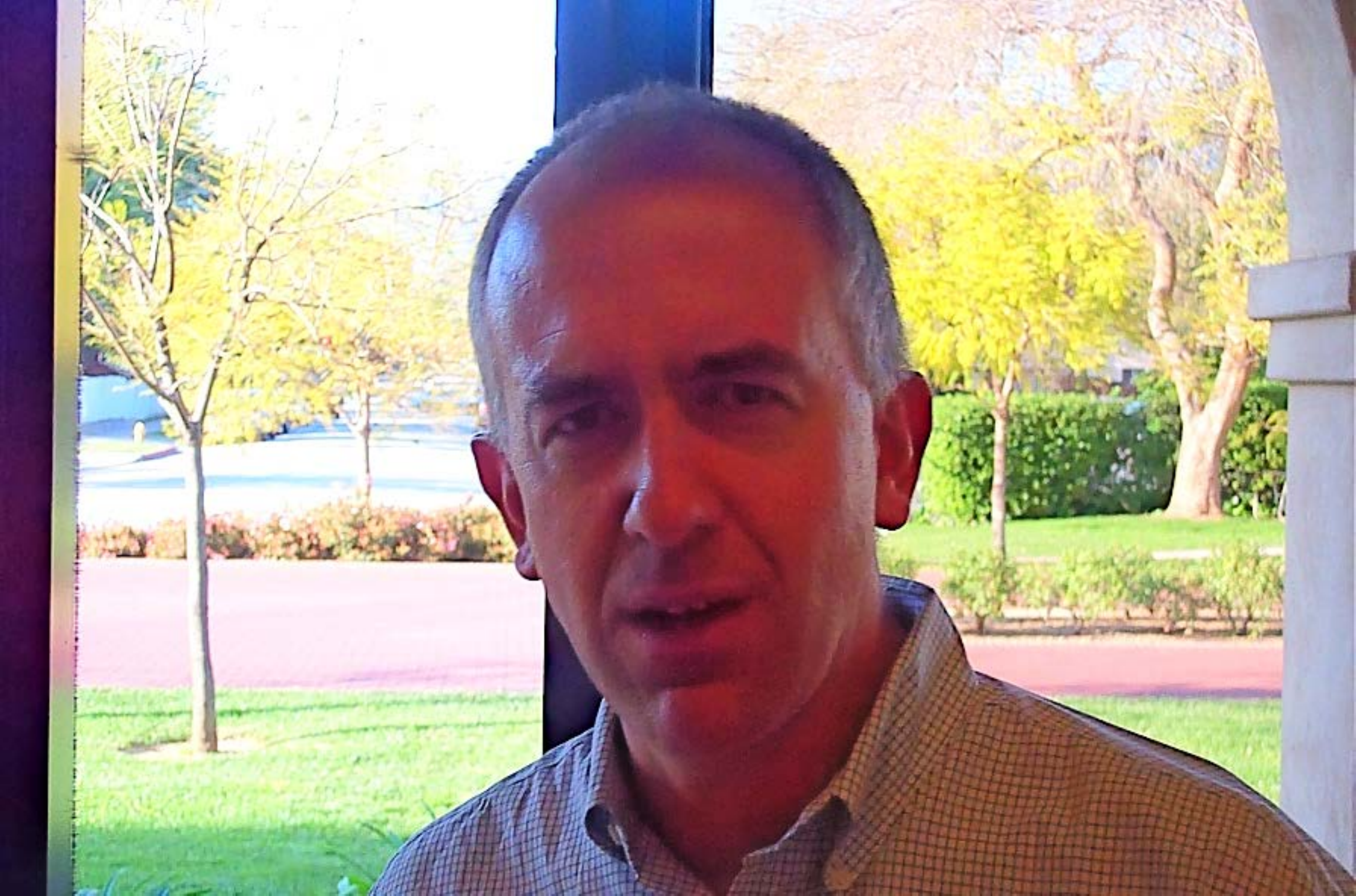} 
            \end{minipage}
        }
        \caption{Comparisons of low-light image enhancement results.}
        \label{F_light}
    \end{figure*}

    \begin{figure*} [ht]
        \centering
        \subfloat[Original]{
            \begin{minipage}[b]{0.15\textwidth}
                \includegraphics[width=1\textwidth,height=1\textwidth]{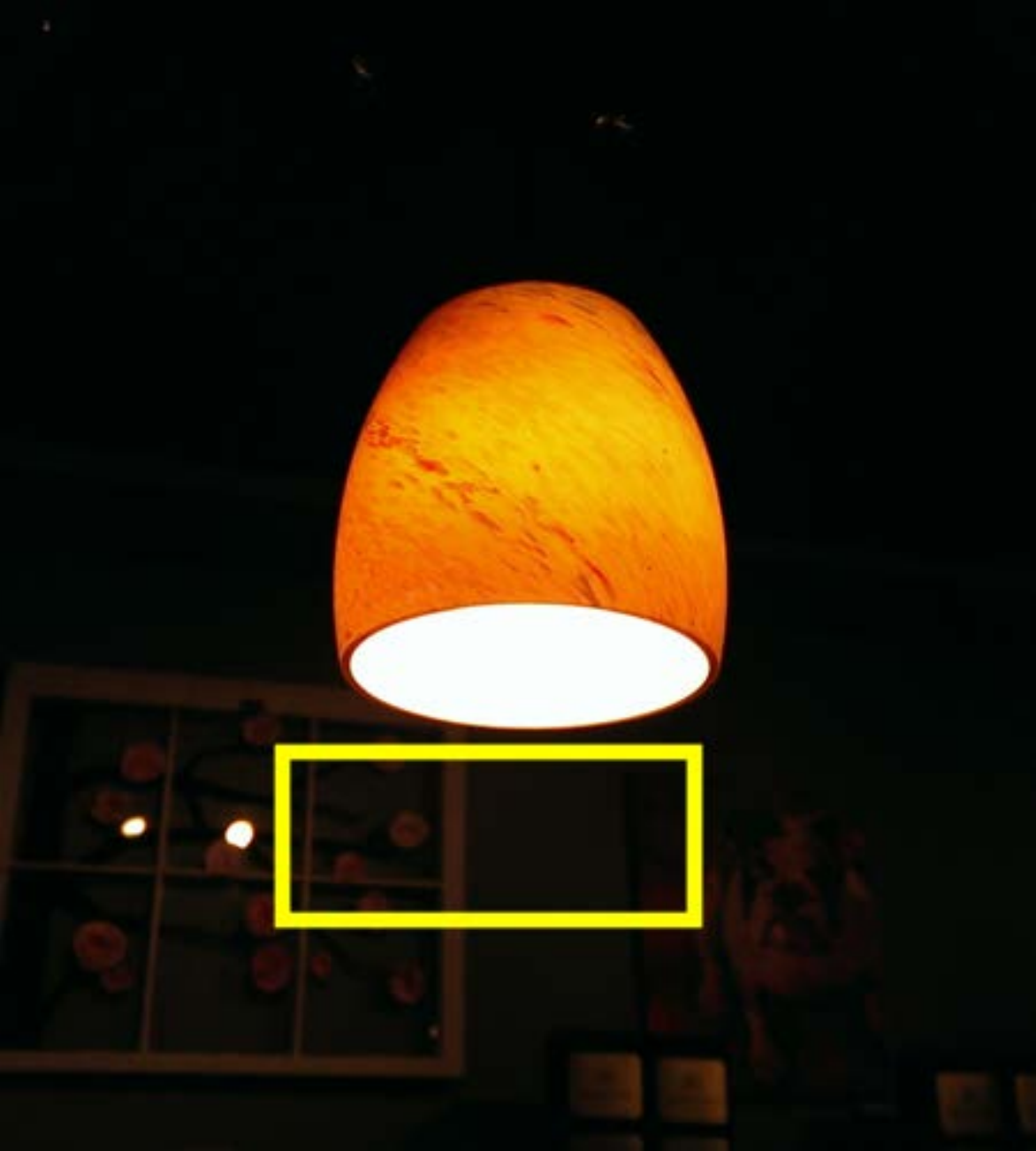} \vspace{-3mm} \\
                \includegraphics[width=1\textwidth]{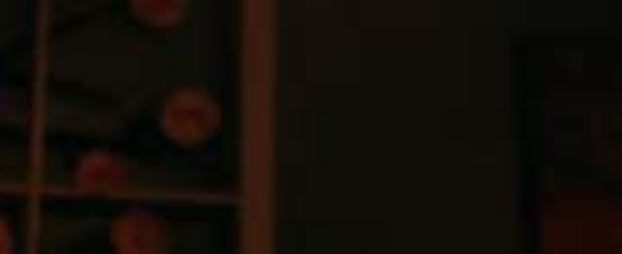} \vspace{-3mm} \\
                \includegraphics[width=1\textwidth]{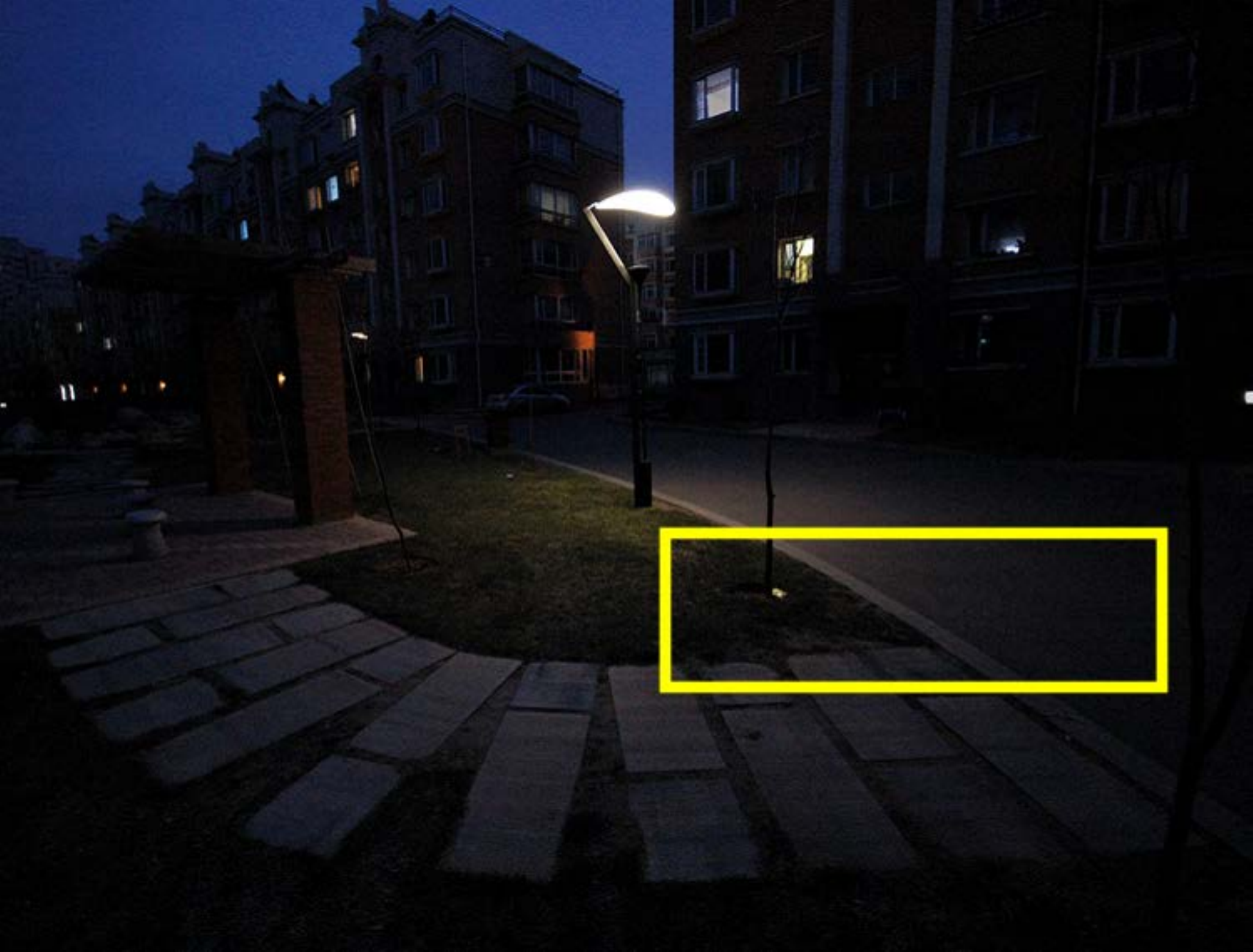} \vspace{-3mm} \\
                \includegraphics[width=1\textwidth]{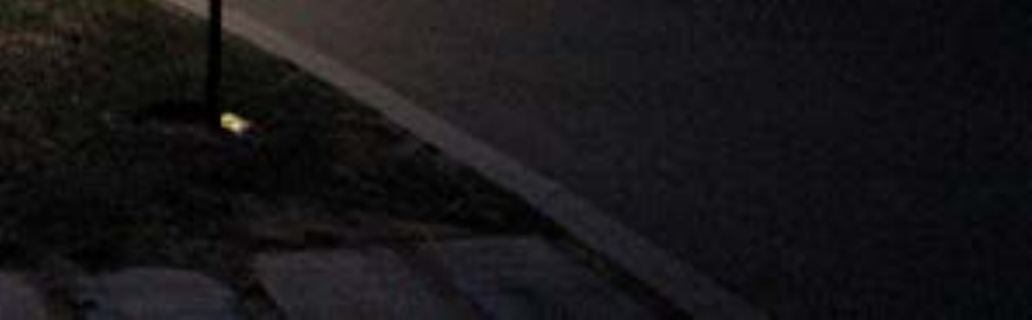}
            \end{minipage}
        } \hspace{-3mm}
        \subfloat[PIE \cite{7229296}]{
            \begin{minipage}[b]{0.15\textwidth}
                \includegraphics[width=1\textwidth,height=1\textwidth]{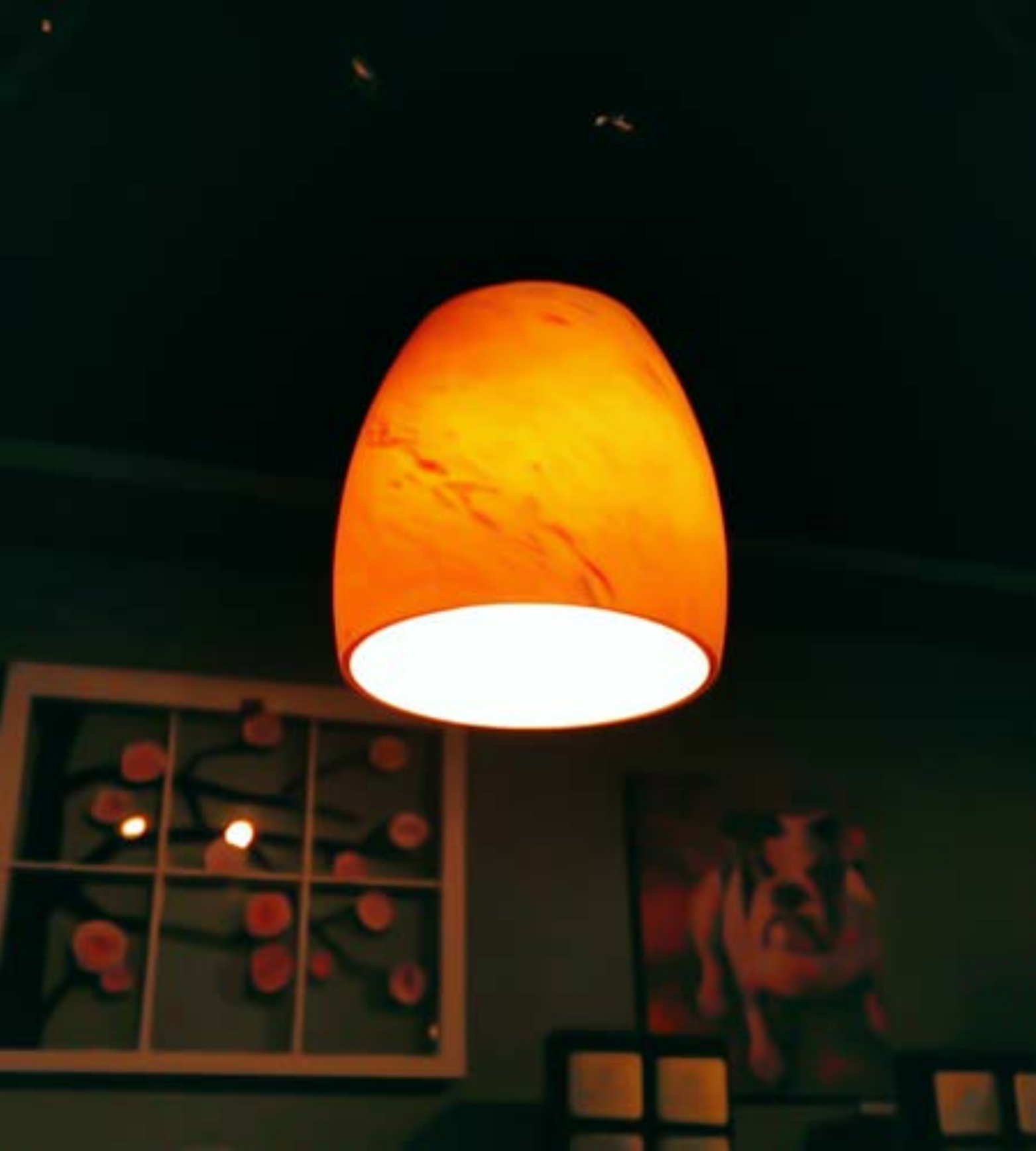} \vspace{-3mm} \\
                \includegraphics[width=1\textwidth]{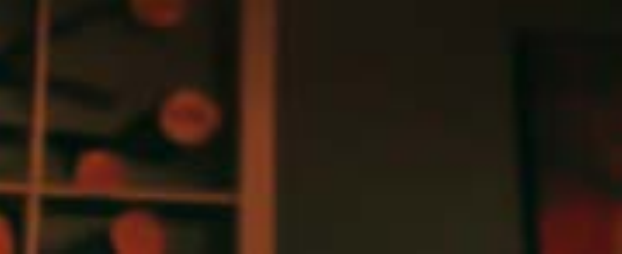} \vspace{-3mm} \\
                \includegraphics[width=1\textwidth]{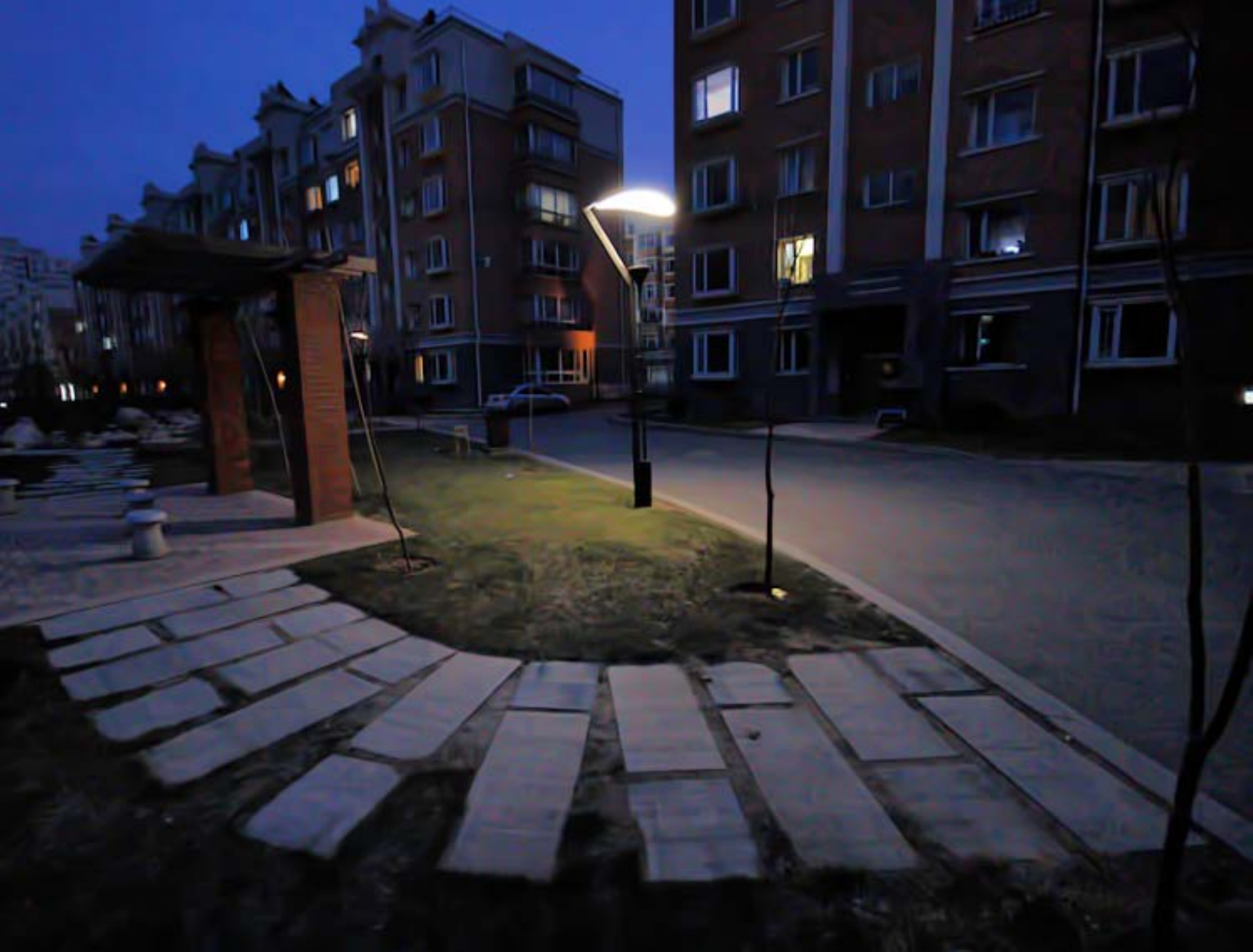} \vspace{-3mm} \\
                \includegraphics[width=1\textwidth]{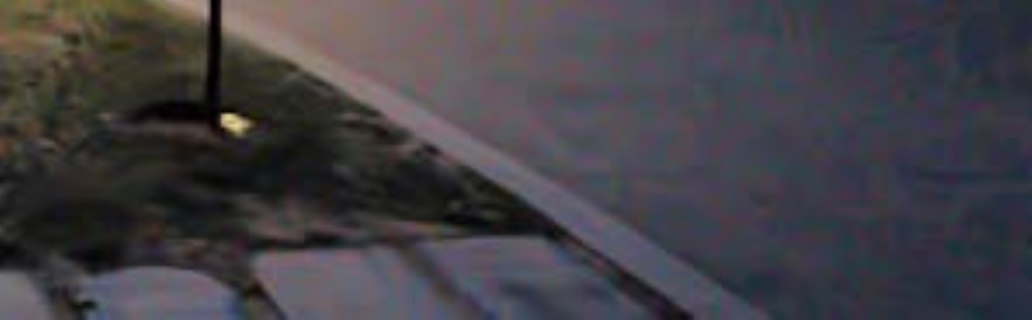}
            \end{minipage}
        } \hspace{-3mm}
        \subfloat[HE]{
            \begin{minipage}[b]{0.15\textwidth}
                \includegraphics[width=1\textwidth,height=1\textwidth]{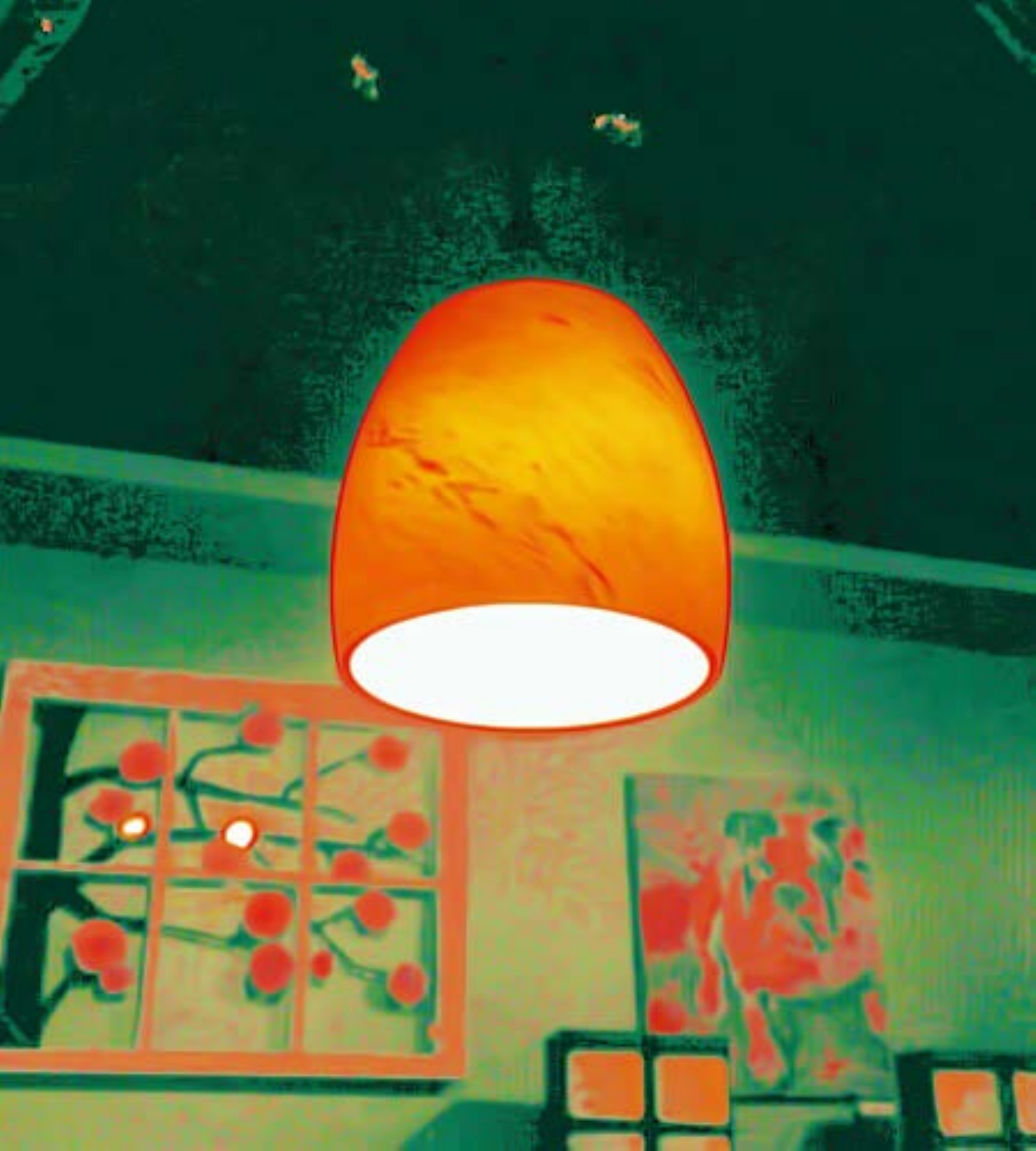} \vspace{-3mm} \\
                \includegraphics[width=1\textwidth]{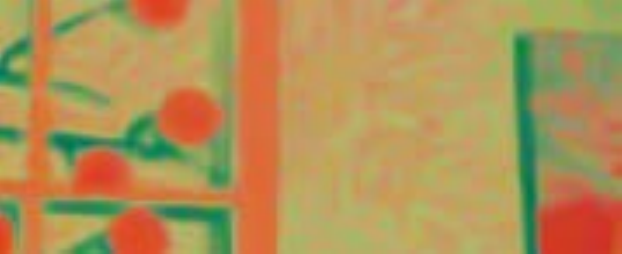} \vspace{-3mm} \\
                \includegraphics[width=1\textwidth]{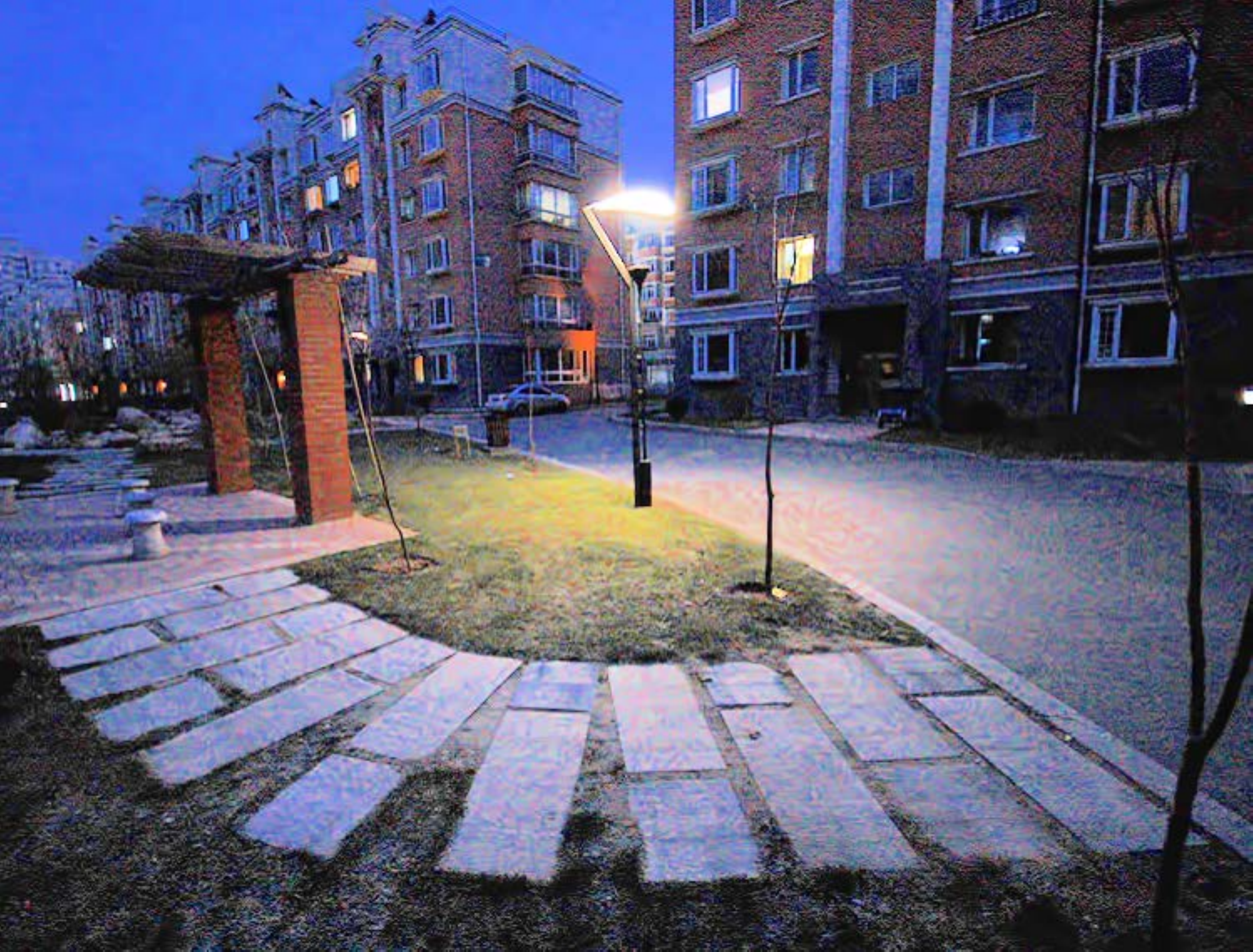} \vspace{-3mm} \\
                \includegraphics[width=1\textwidth]{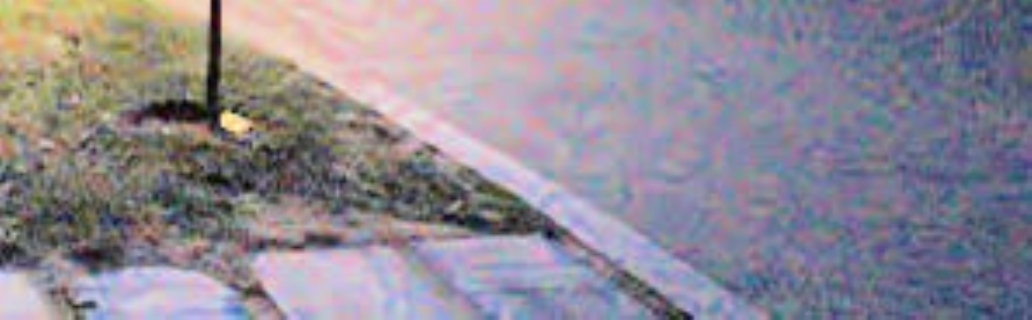}
            \end{minipage}
        } \hspace{-3mm}
        \subfloat[LIME \cite{7782813}]{
            \begin{minipage}[b]{0.15\textwidth}
                \includegraphics[width=1\textwidth,height=1\textwidth]{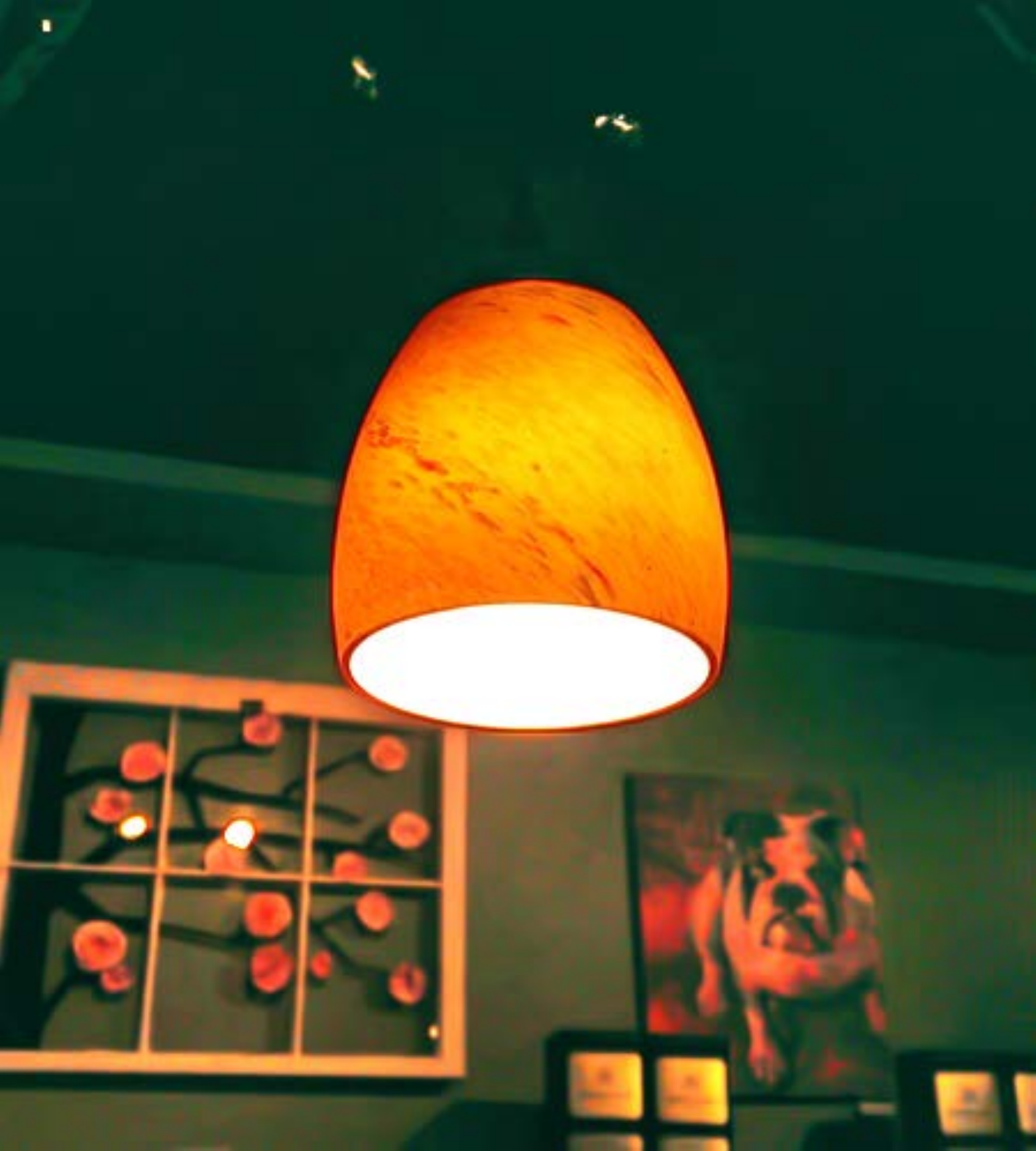} \vspace{-3mm} \\
                \includegraphics[width=1\textwidth]{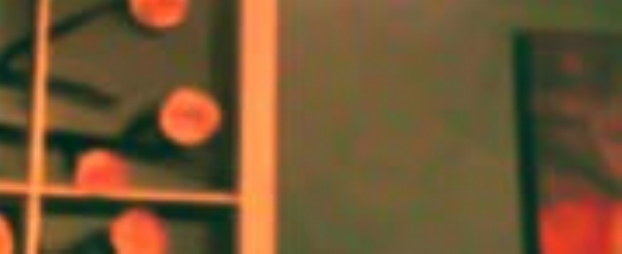} \vspace{-3mm} \\
                \includegraphics[width=1\textwidth]{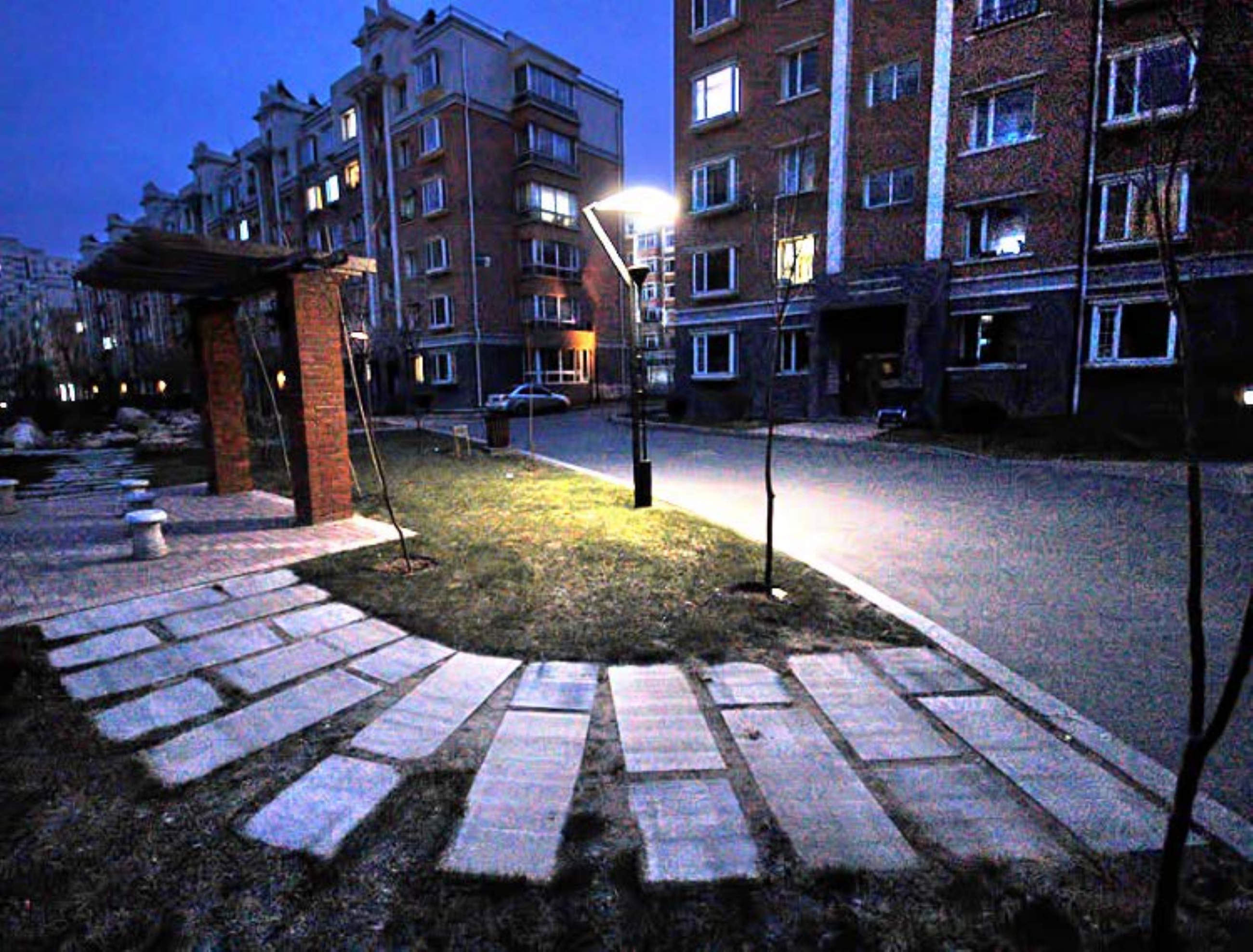} \vspace{-3mm} \\
                \includegraphics[width=1\textwidth]{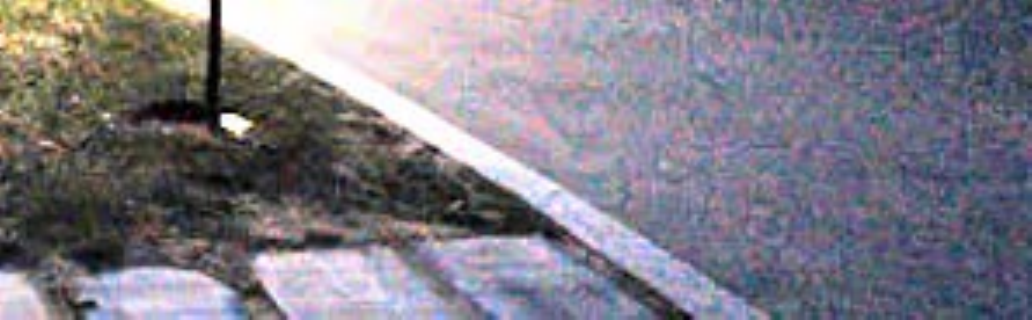}
            \end{minipage}
        } \hspace{-3mm}
        \subfloat[NPEA \cite{6512558}]{
            \begin{minipage}[b]{0.15\textwidth}
                \includegraphics[width=1\textwidth,height=1\textwidth]{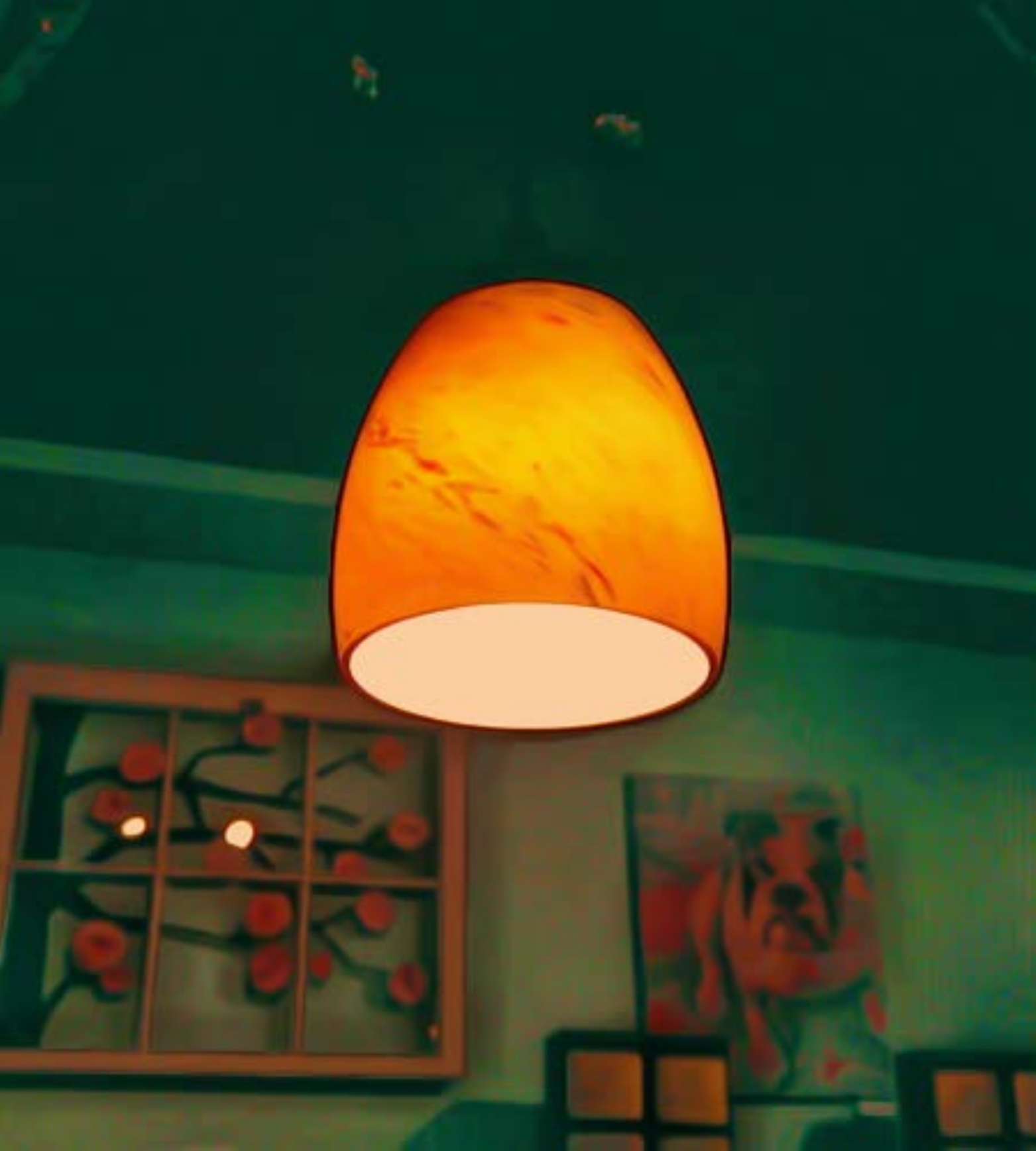} \vspace{-3mm} \\
                \includegraphics[width=1\textwidth]{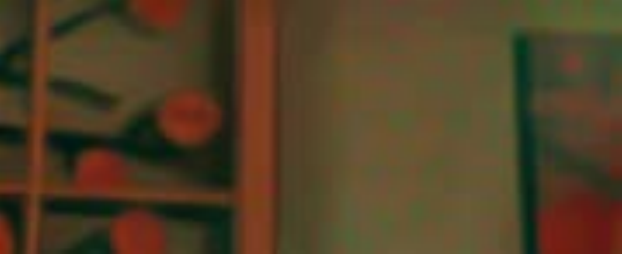} \vspace{-3mm} \\
                \includegraphics[width=1\textwidth]{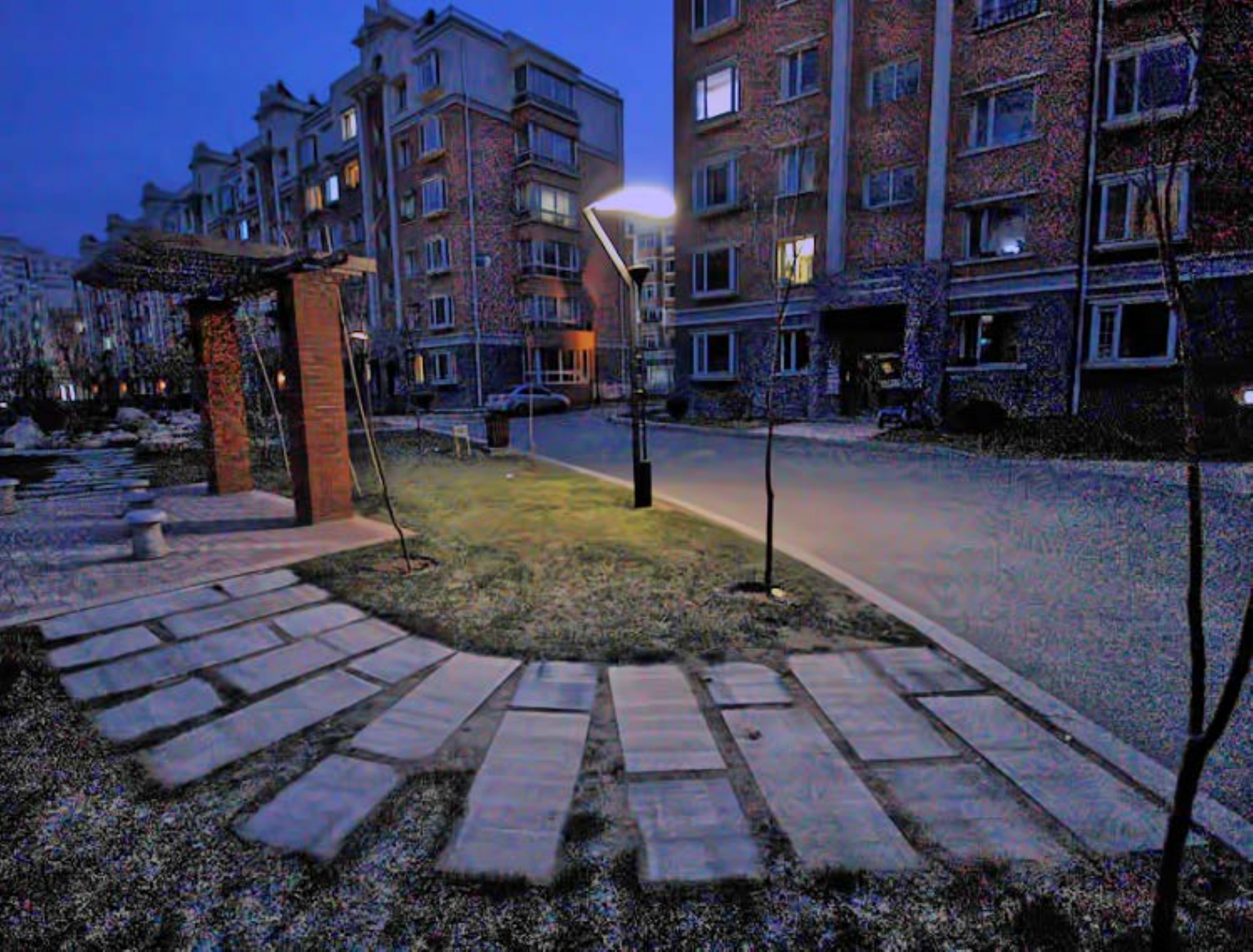} \vspace{-3mm} \\
                \includegraphics[width=1\textwidth]{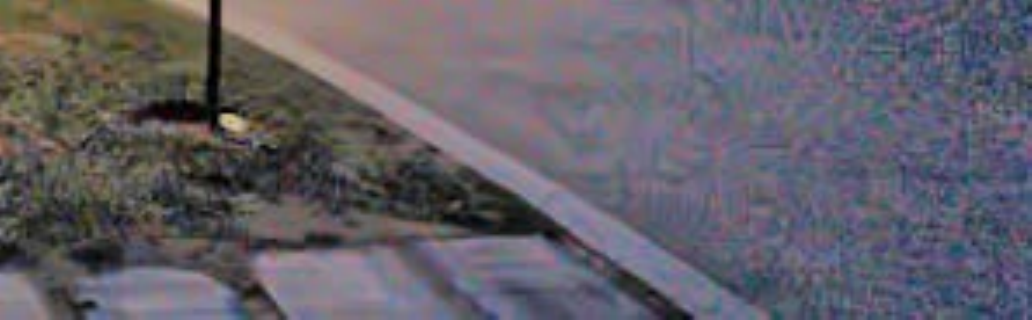}
            \end{minipage}
        } \hspace{-3mm}
        \subfloat[Our Method]{
            \begin{minipage}[b]{0.15\textwidth}
                \includegraphics[width=1\textwidth,height=1\textwidth]{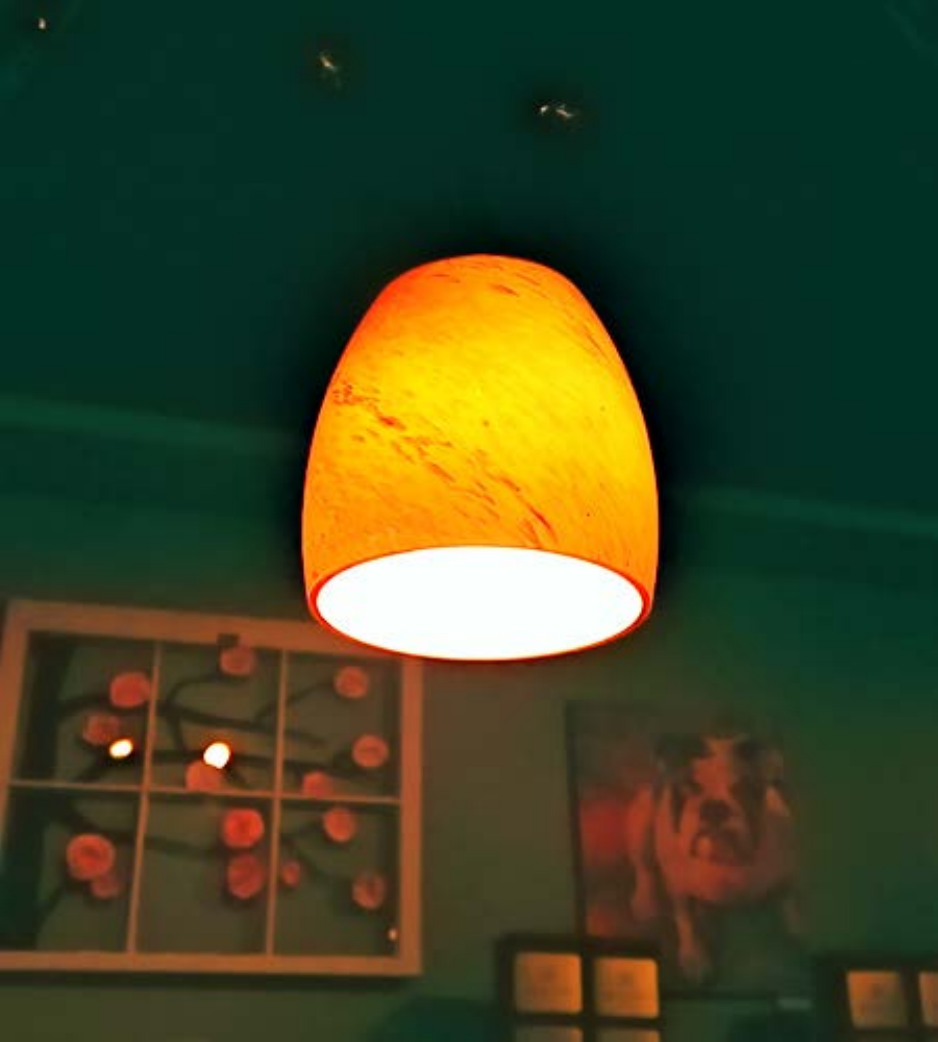} \vspace{-3mm} \\
                \includegraphics[width=1\textwidth]{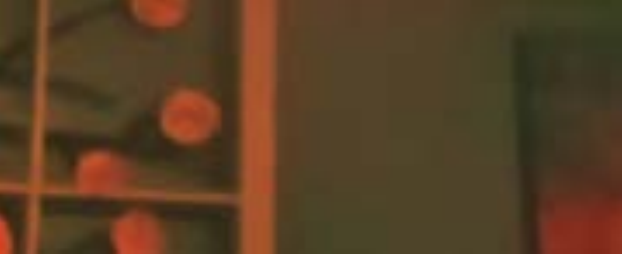} \vspace{-3mm} \\
                \includegraphics[width=1\textwidth]{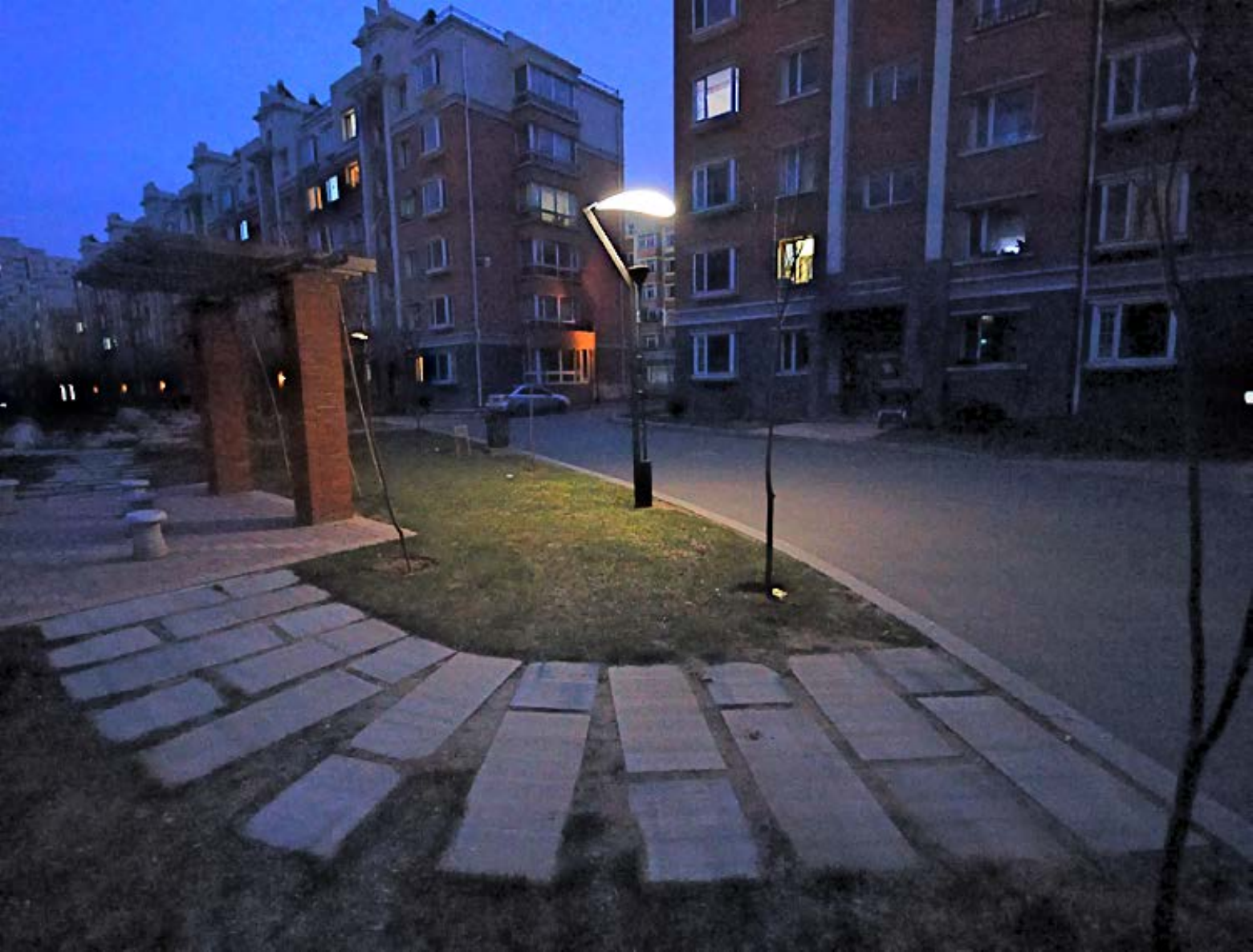} \vspace{-3mm} \\
                \includegraphics[width=1\textwidth]{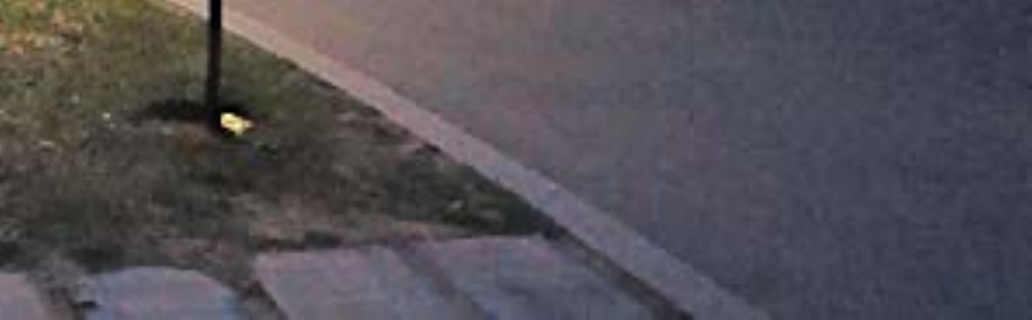}
            \end{minipage}
        }
        \caption{Comparisons of noisy low-light image enhancement results, all methods except ours are followed by BM3D \cite{4271520}.}
        \label{F_noise}
    \end{figure*}

    After the estimation of the illumination $L$ and the reflectance $R$, the gamma correction operation is applied in order to adjust the illumination. And the final enhancement result $S'$ is generated by gamma correction as $S' = R\circ L'^{\frac{1}{\gamma}}$, where $L'$ is the normalized $L$, and $\gamma$ is empirically set as 2.2.

\section{Experimental Results}

    All experiments are performed on MATLAB R2017a with 4G RAM and Intel Core i5-4210H CPU @2.90GHz. In our experiment the parameters $\alpha$, $\beta$ and $\omega$ in (\ref{E_L}) and (\ref{E_R})are empirically set as 0.007, 0.001 and 0.016. $\varepsilon$ and $\sigma$ are set to be 10 and $\lambda$ is set to be 6. In general cases, this setting performs pleasant outcome. Our test images come from dataset provided by authors of \cite{7780673} and \cite{7782813}. More experimental results and code can be found in the authors' webpage \footnote{\url{https://github.com/tonghelen/JED-Method}} .

    First, we compare the illumination, reflectance and outcome images of our method and LIME \cite{7782813} with details. From Fig. \ref{F_decmp} we can see that our illumination maps contain more details than LIME and our reflectance maps generally have more color information. As can be seen in Fig. \ref{F_decmp} (e), our results successfully reserve the local details but erase the noise.

    To evaluate the enhancing effectiveness of our proposed method, we compare it with conventional histogram equalization (HE) and state-of-the-art Retinex based enhancement methods,~\emph{i.e.}~simultaneous reflectance and illumination estimation (SRIE) \cite{7780673}, naturalness preserved enhancement algorithm (NPEA) \cite{6512558}, and LIME \cite{7782813}. From Fig. \ref{F_light} we can find that HE, SRIE and NEPA do not obviously enhance the image and have lots of noise. LIME often over-enhances the image and therefore loses some details in the bright area. Only our method keeps the details of grass in the first image.

    To demonstrate the denoising effectiveness of our proposed method, we compare our results with the results of HE, LIME \cite{7782813}, NPEA \cite{6512558} and probabilistic method for image enhancement (PIE) \cite{7229296}. All methods except our proposed method are followed by an extra denoising procedure via BM3D \cite{4271520}. We can find from Fig. \ref{F_noise} that those enhancement works either make their denoising procedure afterwards difficult and less effective, or often come up with detail-loosing and blurring problems like PIE. Compared to these methods, our method shows strong advantages in both low-light enhancement and denoising.

\section{Conclusion}

    In this paper we discuss the existing problem of noise in mainstream methods of low-light enhancement domain. And we argue that existing methods either ignore this issue or do not handle it well. According to that, we constructively present a joint low-light enhancement and denoising method based on sequential decomposition method. By intentionally limiting noise to the minimum, we can obtain high-quality images finally. Extensive experimental results demonstrate the effectiveness of our method.

\bibliographystyle{IEEEtran}
\bibliography{IEEEabrv,My_Ref}

\end{document}